\documentclass[10pt,journal]{IEEEtran}

\usepackage{hyperref}
\usepackage{url}
\usepackage{amsmath}
\usepackage{bm}
\usepackage{amsfonts}
\usepackage{algorithm}
\usepackage{algorithmic}
\usepackage{amsthm,paralist}
\usepackage{enumitem,graphicx}
\usepackage{color}
\usepackage{subfigure}
\usepackage[normalem]{ulem}
\usepackage{comment}

\usepackage{setspace}

\usepackage[scriptsize,bf]{caption}
\usepackage{amssymb}

\newcommand{\beq}{\begin{equation}}
\newcommand{\eeq}{\end{equation}}

\newcommand{\requ}{\text{ReQU}}

\newtheorem{theorem}{Theorem}[section]

\newtheorem{proposition}{Proposition}[section]
\newtheorem{definition}{Definition}[section]
\newtheorem{claim}{Claim}[section]

\newtheorem{conjecture}{Conjecture}[section]

\newif\ifarxiv

 \arxivtrue

\title{The  Global Landscape of Neural Networks: An Overview }

\author{
	Ruoyu Sun,
	\and 
	Dawei Li,
	\and 
Shiyu Liang,
	Tian Ding,
	R Srikant
	\thanks{
	Author affiliations:	Ruoyu Sun (\texttt{ruoyus@illinois.edu}), Dawei Li (\texttt{dawei2@illinois.edu}),
		Shiyu Liang (\texttt{sliang26@illinois.edu}) and R. Srikant (\texttt{rsrikant@illinois.edu})  are with Coordinated Science Laboratory, University of Illinois at Urbana-Champaign, Urbana, IL.
		Tian Ding  (\texttt{tianding@link.cuhk.edu.hk}) is with the Department of Information Engineering , the Chinese University of Hong Kong.
	   }
}

\begin{document}

\maketitle

\vspace{-0.5cm}
\begin{abstract}

 One of the major concerns for  neural network training is that the non-convexity of the associated loss functions  may cause bad landscape. The recent success of neural networks suggests that their loss  landscape  is not too bad, but what specific results do we know about the landscape?  In this article, we review recent findings and results on the global landscape of neural networks. 
  First, we point out that wide neural nets may have sub-optimal local minima under certain assumptions. 
  Second, we discuss a few rigorous results on the geometric properties of wide networks such as ``no bad basin'', and some modifications that eliminate sub-optimal local minima and/or   decreasing paths to infinity. 
 Third, we discuss visualization and empirical explorations of the landscape for practical neural nets.  Finally, we briefly discuss some convergence results and  their relation to landscape results.

\end{abstract}

\vspace{-0.3cm}
\section{Introduction}
\vspace{-0.3cm}

Deep neural networks have led to remarkable empirical successes in various artificial intelligence tasks, sparking an interest in the theory behind their architectures and training. 
In the early days when the power of neural networks were not fully harnessed, researchers favored models such as supporting vector machines which could be studied using convex optimization techniques. A major concern in the case of neural networks is that the non-convexity of the associated loss functions
may cause complicated and strange optimization landscapes. 
However, recent experience shows that neural networks can often be trained to find the global minima of appropriately chosen loss functions, thus  it is  of great interest to understand the loss landscape of neural networks. 

A closely related problem is to understand the landscape of the objectives in non-convex matrix problems.
In this context, it has been established that the landscape is benign (e.g., every local minimum is a global minimum) for quite a few matrix problems, such as matrix completion and phase retrieval, under certain assumptions (see, e.g., \cite{chi2019nonconvex} for a survey).
Although there are still many cases that remain difficult to analyze, there is much optimism that non-convex matrix problems (under proper assumptions) often have benign landscapes.

For the landscape of neural networks, the status is less clear. One might be interested in getting a ``yes'' or ``no'' answer to questions such as  ``does a neural network have sub-optimal local minima?'', or ``can a neural-net problem be solved to find global minima?''. 
Much progress has been made, but
 a fully satisfying answer is still not available.  
We hope this article can explain the existing results
in a coherent way so that they are
relatively easy to understand.

Compared to another recent survey
\cite{sun2019}, this article
focuses on global landscape and contains formal
theorem statements, while \cite{sun2019}
covered many aspects of neural net optimization
and did not present formal theorems. 
We suggest readers unfamiliar with neural-net optimization
to read \cite{sun2019} for a big picture,
and read this article for
more in-depth understanding of the global landscape.
We focus on results that can apply to deep
nets, thus we do not discuss
 many results on shallow nets which
 are reviewed in \cite{sun2019} (there
 is a restriction on the number of references
from the journal, so we only select
  a subset of references in this article).

  \subsection{Summary}

The goal of this survey is to provide an overview of the recent progress on the  global landscape of neural networks.
The central questions to answer are: 
\begin{itemize}
    \item Q1: 
    How to \textit{explain} the good performance of neural-net optimization algorithms, despite
    the non-convexity?

    \item Q2 (rigorous evidence for Q1): 
    To prove a rigorous result, what conditions are needed, and what property can be established?
    
    \item Q3: How to \textit{design} the system so that
  a rigorous theory can be established? 
  
\end{itemize}

Based on these high-level questions, we organize the existing results in a flow as follows. 

\begin{itemize}
	\item 
	An initial explanation is
	Hypothesis 1: ``every local-min is a global-min''. 
	The first rigorous evidence is that
	deep linear networks have no sub-optimal local minima, under mild conditions (Sec. \ref{sec: linear}).
	
	\item However, practitioners find that 
	  narrow neural-nets cannot be solved well,
	  while over-parameterized neural-nets can. 
	  Thus researchers believe that a crucial  condition is ``over-parameterization''. Results on linear networks do not utilize this condition, thus are not enough to explain practice.
	
	\item  %
	For non-linear over-parameterized neural nets, sub-optimal local-min can exist under certain
	assumptions (Sec. \ref{subsec: counter-example}).
	An alternative Hypothesis 2 is that
local descent algorithms ``avoid'' sub-optimal local-min in neural-net training. %

	\item %
One explanation of Hypothesis 2 is the following.
For many over-parameterized neural-nets no ``bad valleys/basins'' exist (Sec. \ref{subsec: absence of valleys}). Thus even if sub-optimal local minima exist, they cannot be strong attractors
and thus iterates will not be attracted to them (see
Sec. \ref{subsec: intution and open questions}
for such a conjecture).

 \item 
 With stronger assumptions (e.g. ultra-wide nets),
 it can be rigorously proved that
 gradient descent can avoid reaching the area with sub-optimal local minima,
 thus converging to global minima
(Sec. \ref{subsec: algorithm convergence}). 
Due to space reason, we only touch the surface of this sub-area in this article.

	\item 
	{ The above research assumes no modification of the
	neural-net landscape.  }
If we are allowed to design the landscape (e.g., adding regularizers), then proving ``every local-min is global-min'' becomes possible for a wide range of neural networks (Sec. \ref{subsec: eliminate local-min}). We
further discuss a result that ensures the absence of both bad local-min and decreasing path to infinity (Sec.
	\ref{subsec: add reg to smooth}).

	\item Finally, what is the lesson for empirical training? Successful training of neural-nets requires proper initialization, batch normalization, residual connection and wide/deep networks (see \cite{sun2019} for a more thorough survey). The current article 
focuses on one empirical lesson: large width is important for successful training. 
For practitioners, many theoretical results in this article can be viewed as evidence of this lesson.
	
\end{itemize}

\subsection{Big picture: the role of landscape analysis}

Landscape analysis has been a subject of study since 1980's; see \cite{bianchini1996optimal} for an overview. 
The concern that gradient descent can get stuck at bad local minima  has been around for a long time. For instance, Minsky and Papert commented in  \emph{Perceptrons} (expanded edition) that ``\textit{they speak as though becoming trapped on local maxima were rarely a serious problem}'' and ``\textit{we conjecture ... will become increasingly intractable as we increase the numbers of input variables}''.
Reference \cite{bianchini1996optimal} argued that
to address Minky and Papert's comment, it is ``\textit{very interesting to investigate the presence of local minima}''. 
Our survey can be viewed as a modern version of the survey \cite{bianchini1996optimal},
by including new results and new understanding, especially
results on deep networks. 
In particular, we will point out that a main claim
reviewed in  \cite{bianchini1996optimal} that ``over-parameterized 1-hidden-layer network have no sub-optimal local minima'' is not rigorous.

The theory of machine learning (for supervised learning) consists of three parts:
representation, optimization and generalization. One way to interpret this partition is via the lens of error decomposition:
the test error can be decomposed as the sum of representation error,  optimization error and generalization error.
Landscape analysis is an important component of understanding the optimization error. 
The optimization error refers to
$ F(\hat{w} ) - F^* $, where $F$ is the loss function, $\hat{w}$ is the solution (i.e, the parameters of the neural network) found by an algorithm,
and  $F^*$ is the the globally minimal loss.
Define $w^{\infty} $ to be a converged solution if the algorithm runs 
for infinite time and assume it converges. 
The optimization error can be further decomposed into two parts: 
$$ [  F(  \hat{w} ) - F(w^{\infty })  ]  +   [    F(w^{\infty})  - F^*   ]   . $$
The first part is the ``non-convergence error'',     %
which occurs because either the algorithm  is  intrinsically not convergent 
or  the algorithm has not converged yet due to limited running time. 
It is often reasonable to assume $w^{\infty} $ is a stationary point or even a local-min.
The second part is the ``infinite-time error'', which indicates how far away the converged value is from the global minima value. 
If every local-min is a global-min, and $w^{\infty} $ is a local minimum,
then this term becomes $0$. Proving the convergence of an algorithm is a central task of classical optimization, but it often does not cover the ``infinite-time error''.
Therefore, landscape analysis provides an understanding of the fundamental limit  of the loss function, and is somewhat similar to Shannon's capacity bound \footnote{We draw this analogy since we expect many readers are from
signal processing and information theory area.}: it indicates how well an algorithm can possibly perform with long training time. 

Although our focus is on landscape analysis, we will briefly discuss how a good landscape can possibly lead to the convergence to global-min.
Another important topic
is the relation of optimization and
generalization, such as implicit regularization
(e.g. \cite{neyshabur2017implicit,li2018algorithmic})
and the conjecture that wide minima generalize
better (e.g. \cite{keskar2016large}).
Due to space, we do not discuss generalization  in this article.

\section{Models}

In this section, we present the optimization formulation for a supervised learning problem. 
Consider input instances  $x_i \in \mathbb{R}^{d_x}$ and output instances
$y_i \in \mathbb{R}^{ d_y } , i=1, \dots, n$, where $n$ is the number of samples. 
The goal is to build a model that can  predict $y_i$ based on  $x_i$.
We use a neural network $ f_{ \theta }:  \mathbb{R}^{d_x}  \rightarrow  \mathbb{R}^{ d_y }  $
to produce a prediction $\hat{y}$ based on an input $x$. 
For most parts of the article, we consider a fully-connected neural network 
\begin{multline}\label{neural net def}
f_{\theta} (x) =  W_{ L  } \phi (   W_{ L-1} \dots \\
\phi(  W_2 \phi (W_1 x  + b_1)  + b_2 ) \dots + b_{L-1} ) ,
\end{multline}
where  
$\phi:  \mathbb{R} \rightarrow  \mathbb{R} $ is the neuron activation function (or simply ``activation''), $W_j $ is a matrix of dimension $d_{j }\times d_{j - 1 } $, $j =  1, \dots, L$ and $\theta = ( W_1, b_1, \dots, W_{L-1},b_{L-1},W_L  )$ is the collection of all parameters. 
Note that we denote $d_0 = d_x$ and $d_L = d_y$.
We will use $\phi(Z)$ to denote a matrix with each entry $ \phi(Z)_{ij} $ being $ \phi(Z_{ij}).$

For a certain loss $ \ell(\cdot, \cdot)  $, the problem of finding the optimal parameters can be written as
\begin{equation}\label{main problem}
\min_{\theta }  F(\theta) \triangleq  \frac{1}{ n } \sum_{i=1}^n  \ell( y_i,  f_{\theta} ( x_i )  ). 
\end{equation}
For regression problems, $\ell(y, z) $ is often the quadratic loss  $\ell(y,z) = \| y - z \|^2$. 
For binary classification problem, a popular choice of $\ell$ is
the logistic loss $\ell (y, z) = \log (1 + \exp ( - y z) ) $.

Finally, we present a few standard definitions. 
We say $ \bar{\theta} $ is a global minimum (or simply global-min) of function $ F $
iff $ F( \bar{\theta}) \leq F(\theta), \forall \; \theta.  $
We say $ \bar{\theta} $ is a critical point of function $ F $
iff $ \nabla F( \bar{\theta} ) = 0. $
We say $ \hat{\theta} $ is a local minimum (or simply local-min) of a function $ F (\theta)$
iff there exists an open set $ B $  that contains $ \hat{\theta} $ such that  $ F( \hat{\theta}  ) \leq F( \theta  ), \forall  \theta \in B $.
We say $ \hat{\theta} $ is a strict local-min if the inequality is strict for any other $\theta \in B$.
The local maximum can be defined in a similar way (replacing $\leq$ by $\geq$). 
We say $ \hat{\theta} $ is a saddle point iff it is a critical point and neither a local minimum nor a local maximum.

\section{Linear neural networks }\label{sec: linear}

The initial hypothesis is that ``every local-min is global-min''
in practical neural-nets. 
The results on linear neural networks were considered
early evidence (though not strong), thus historically important.
Besides the historical reasons, studying linear networks 
can help develop technical tools. We remark that linear neural networks are rarely used in practice since their representation power is the same as a linear model, so non-theory readers can skip this section if not interested.

\begin{figure}
	\begin{minipage}[t]{0.5\linewidth}
		\centering
		\includegraphics[width=0.9\linewidth]{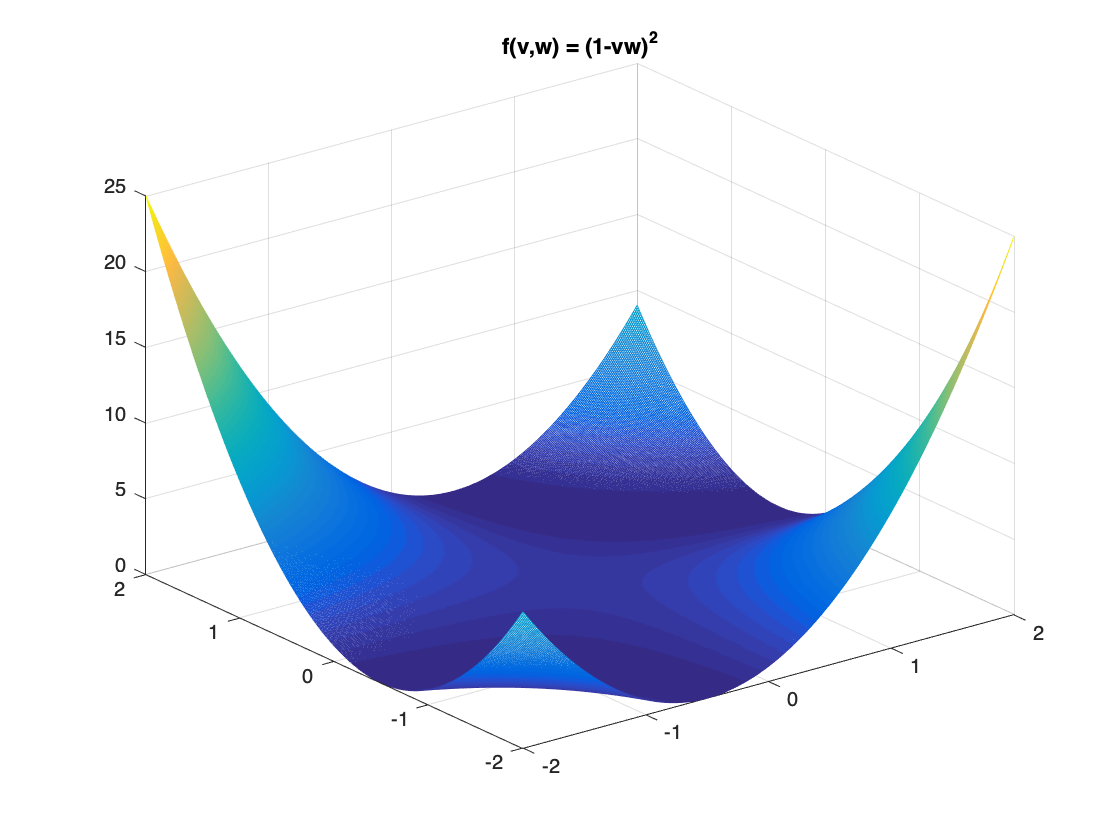}
		\label{fig:side:a}
	\end{minipage}%
	\begin{minipage}[t]{0.5\linewidth}
		\centering
		\includegraphics[width=0.9\linewidth]{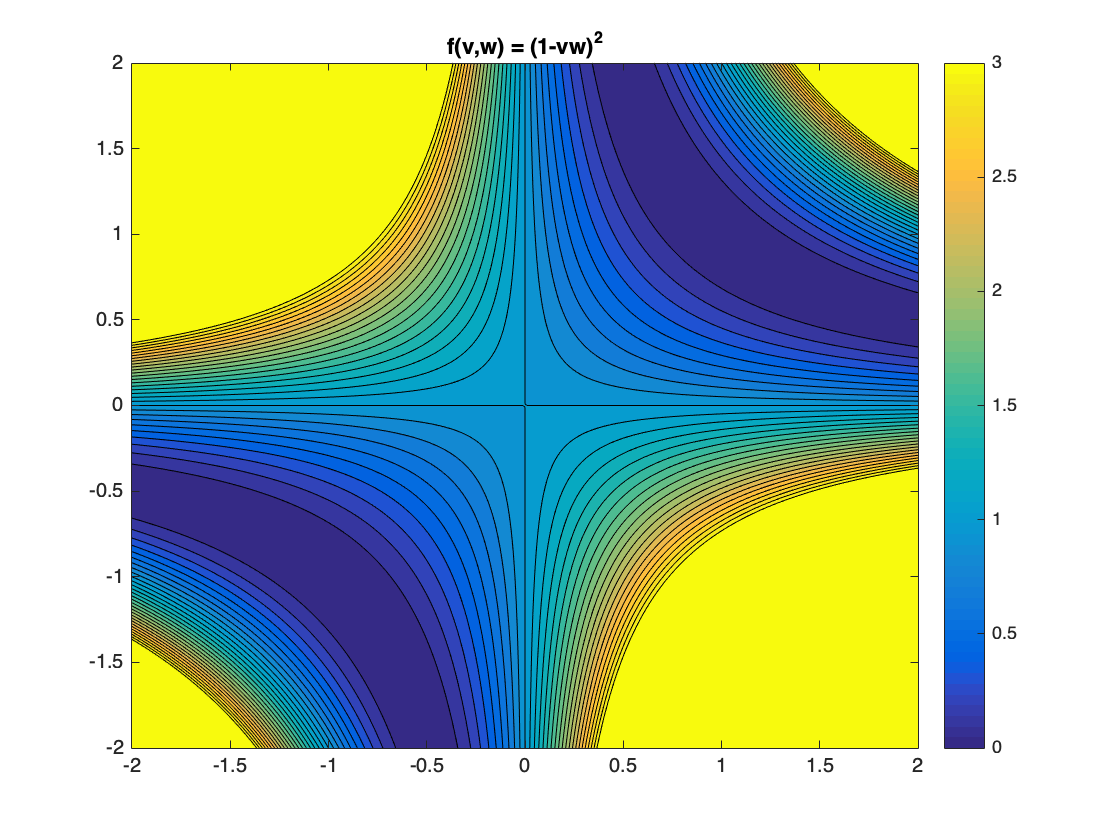}
		\label{fig:side:b}
	\end{minipage}%
	\caption{The loss surface of $f(v, w)=(1-vw)^2$. 
		All global minima lie in the two curved regions in dark blue. }
	\label{fig:linear}
\end{figure}

\textbf{Toy example}. 
We consider the simplest linear neural network problem
$$
\min_{ v, w \in \mathbb{R} }  (vw - 1 )^2.  
$$
This is a non-convex problem, but it is easy to prove that
every local minimum  is a global minimum. 
We plot the function and its contour in Figure
\ref{fig:linear}.

\textbf{Hamiltonian of a spin-glass system}. 
Choromanska et al. \cite{choromanska2015loss} analyzed the global landscape
of multi-layer networks. The motivation was to study a multi-layer network with ReLU activations,
but the ReLU activations are removed by adding a somewhat unrealistic assumption,
thus essentially converting the network into a multi-layer
linear network. 
Under a few other assumptions, the loss function 
is transformed to a polynomial function 
$ \sum_{ i_1, \dots, i_L  = 1 }^p   X_{i_1, i_2, \dots, i_L } w_{i_1} \dots w_{i_L} $
with Gaussian random coefficients $X_{i_1, i_2, \dots, i_L } $.
\begin{definition}(index)
The index of a critical point is the number of negative eigenvalues of the Hessian at this point. 
\end{definition}
They  computed the limit of  the expected number of stationary points
with a given index  as the width $p $ goes to infinity. 
 Based on the calculations, they described a layered structure for stationary points with different indices: low-index stationary points (including local minima) are closer to global minima than high-index stationary points (the precise statement is highly technical and omitted here). 
 While their neural network model is somewhat far from practice, the description of the landscape is rather unique and not seen in other works
 \footnote{We remark that there might be a trade-off between intuition and rigor: \cite{choromanska2015loss} covers not only local-min but also other critical points, thus contains ``more intuition'' than works that only study  local-min; meanwhile, the result requires more unrealistic assumptions than other works as well. A reader may find this result more interesting or less interesting, depending on how much rigor they expect. }.

\textbf{Deep linear networks}. 
Kawaguchi \cite{kawaguchi2016deep} extended an early work  \cite{baldi1989neural}
on 2-layer linear networks to deep linear networks, showing that  every local minimum is a global minimum.
More specifically, the following problem was studied:
\begin{equation}\label{form: deep linear}
\text{P}_1: \quad    \min_{ W_1, W_2, \dots, W_L } \| Y - W_L W_{L-1} \dots W_1 X \|_F^2,
\end{equation}
where $W_i \in \mathbb{R}^{d_{i} \times d_{i-1} } , i=1, \dots, L$.
\cite[Theorem 2.3]{kawaguchi2016deep} is the first landscape result
 on this problem; below we state a slightly stronger version in \cite{lu2017depth}.
\begin{theorem}\cite{lu2017depth}
	\label{thm::linear_no_bad_localmin}
	Suppose $X$ and $Y$ have full row rank,
	then every local minimum of the problem \eqref{form: deep linear} 
	is a global minimum. 
\end{theorem}

The proof of \cite{kawaguchi2016deep} is rather complicated, and \cite{lu2017depth} provided a simpler and more intuitive proof.
The idea of \cite{lu2017depth} is to view the optimization problem \eqref{form: deep linear}
as a re-parameterization of a ``mother'' problem $\text{P}_2:$
$\min_{ \text{rank}( R ) \leq  p } \| Y - R X \|_F^2  $ where
$p = \min \{  d_0, d_1, \dots, d_L \} $.
Note that the effective search space of the original problem
  $ \{  W_L \dots W_2 W_1 \mid  W_i \in \mathbb{R}^{d_{i} \times d_{i-1} }  \} $
   is the same as the  new search space 
     $ \{  R: \text{rank}( R ) \leq  p   \}$, which is why we call P2 a ``mother problem''.
The first step is to  prove that any local-min of $P_1$ achieves
the value of a local-min of $ \text{P}_2 $, and the second step is to prove that $ \text{P}_2 $ has no sub-optimal local-min.

\ifarxiv
\textbf{Characterization of all critical points.}
\cite{yun2017global}   and \cite{zhou2017critical}
provided a more precise characterization of the critical points of deep linear networks.
We briefly discuss the results  of \cite{yun2017global}  for the problem \eqref{form: deep linear}. 
\begin{theorem}\cite{yun2017global}
	\label{thm::linear critical points}
	Assume $d_x \leq n, d_y \leq n$, $XX^T$ and $XY^T$ are full rank,
	$YX^T (XX^T)^{-1} X  $ has distinct singular values. 
	Further, assume  the thinnest layer is the input or the output layer, i.e.,
\begin{equation}\label{thinnest layer}
	 \min_{ 0 \leq i \leq L } d_i  = \min \{ d_x, d_y  \}. 
\end{equation}
	For the problem \eqref{form: deep linear},  every critical point $(W_L, \dots, W_1)$ with the product $W_L  \dots W_2 W_1$ being full rank is a global minimum, and every critical point with $W_L  \dots W_2 W_1$ being singular is a saddle point.
\end{theorem}

The result listed above has a rather clean conclusion: it links
full-rankness to global minima. \textit{Full-rankness}  will appear in the analysis of non-linear networks too, which we discuss later. 
\fi

\section{Over-parameterized Networks}

A major goal of theoretical research is to identify critical factors of
 modern neural networks that contribute to successful training. 
 Nowadays, it is commonly believed that  \textit{large width} is one such factor. 
One evidence is the empirical observation that wide networks are easier
to train than narrow networks (e.g. achieving smaller training and test error).
Another evidence is that pruned models can achieve similar performance to the original model (e.g., \cite{han2015deep}),
implying that there are many redundant parameters to help optimization.
It is thus an interesting theoretical question whether over-parameterization 
indeed leads to a benign landscape. 

 In this section, we discuss the landscape of deep over-parameterized networks (more precisely, wide networks).

\subsection{Toy example: single neuron}

\begin{figure}
	\centering
	\includegraphics[width=6.5cm,height=3.5cm]{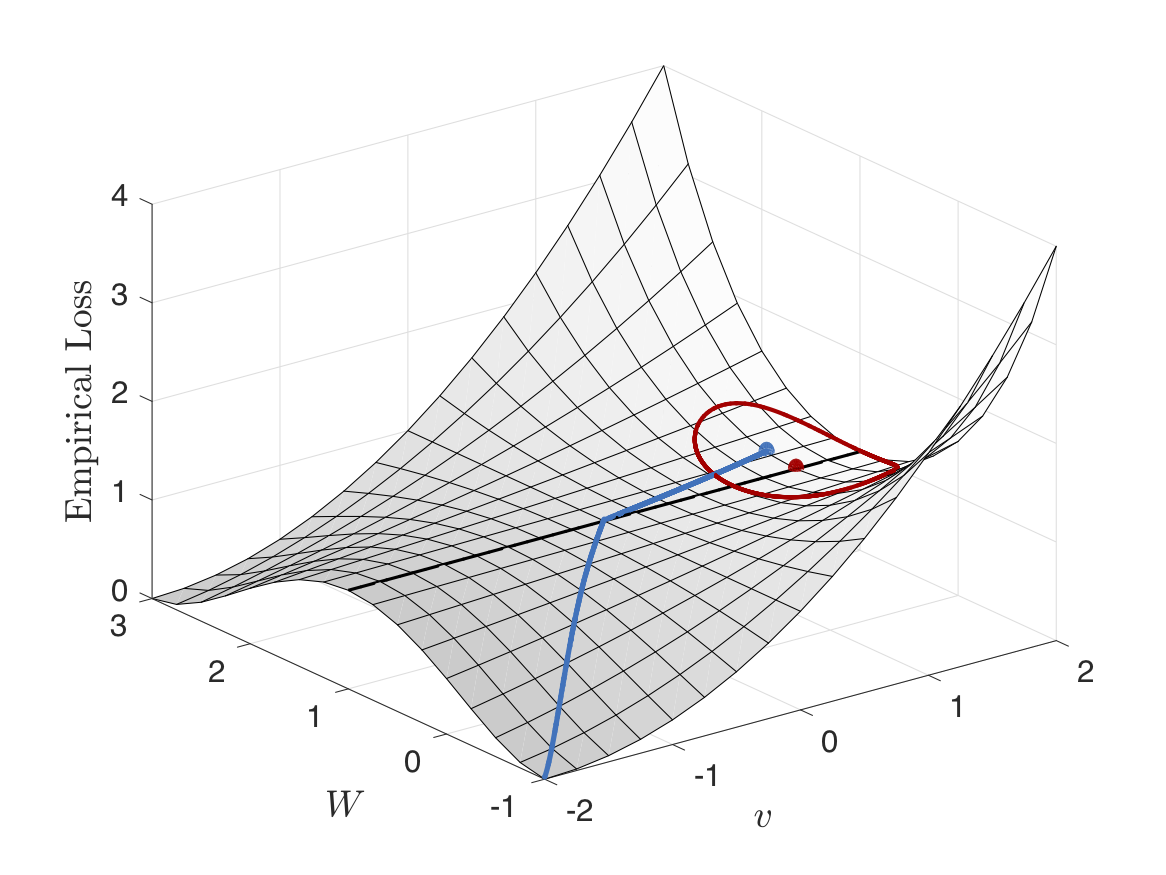}
	\caption{	The loss surface of  $ \min_{ v, w \in \mathbb{R} }  (  y-  v \phi (w x) )^2$
		with  $(x,y) = (1,1)$, and $\phi(z) = -(z-1)^2/8$. The red point $(v,W) = (1,1)$  is a 
		sub-optimal local minimum, but it is not a strict local-min (flat along one
		direction, and curved along the orthogonal direction). }
	\label{fig4_local_min_nonlinear}
\end{figure}

As a toy example, we consider the case $n = d_1 = 1$, 
i.e., a single sample and a non-linear network with a single neuron.
Suppose the associated objective function is 
$
  \min_{ v, w \in \mathbb{R} }  (   1 -  v \phi (w) )^2. 
$
We visualize the landscape  for a special activation  %
in Figure \ref{fig4_local_min_nonlinear}, and it shows that there are infinitely many sub-optimal local minima. 

The following result describes the landscape %
of the above toy model %
for general activation functions.

\begin{proposition}\cite{ding2019spurious}
  \label{thm::bad_localmin_1neuron}
  Suppose $x , y \in \mathbb{R} \backslash \{ 0 \}$.
The function $ F(w,v) = (  y -  v \phi (w x) )^2$ where $v, w \in \mathbb{R}$
 has no sub-optimal local minima if and only if the following condition holds:  if $\phi(t) = 0$,  then $t$ is not a local minimum or local maximum of $\phi$.
\end{proposition}

This result shows that  %
the landscape depends on the neuron activation. For instance, if $\phi(t) = \max \{ t, 0 \}$ (ReLU activation) or $\phi(t) = t^2$, then sub-optimal local-min exists; if $\phi(t ) = t^2 + 1 $ or $\phi$ is strictly increasing, then there is no sub-optimal local-min. When $x$ and $y$ are in high-dimensional space, what conditions guarantee the non-existence of sub-optimal local minima are still not fully understood, though
partial progress has been made. In the next two subsections,
we discuss a few results for more general neural-nets. 

\subsection{Bad local-min for ReLU networks}

Due to the popularity of ReLU activation, a few works analyzed 2-layer ReLU networks. 
For instance,  \cite{liang2018understanding} \cite{safran2017spurious} 
constructed sub-optimal local minima for 2-layer ReLU networks
under different settings. 
The existence of sub-optimal local minima for ReLU networks is not surprising, since Proposition \ref{thm::bad_localmin_1neuron} showed that even for single-neuron network with ReLU activation, sub-optimal local-min can exist.
Nevertheless, a rigorous analysis for multi-neuron
ReLU networks is non-trivial and requires other techniques.
Due to space reason, we do not review these results in detail here.

\subsection{Does over-parameterization eliminate bad local-min, for smooth neurons?}\label{subsec: counter-example}

One major  result reviewed in the survey \cite{bianchini1996optimal}
is that a wide one-hidden-layer network has no sub-optimal local minima. At that time, researchers thought the assumption of ``many hidden neurons'' is restricted. Nowadays, this assumption is considered rather reasonable,
thus it is worthwhile to revisit this classical result more carefully. 
\cite{bianchini1996optimal} did not cite the full result, and we cite the version of \cite{yu95} below. 

  \begin{claim}\cite[Theorem 3]{yu95}
	\label{thm::Yu no local min}
Consider the problem $ \min_{ v \in \mathbb{R}^{1 \times p},  W  \in \mathbb{R}^{p \times d_x }} \sum_{i = 1}^n  (y_i -  v \phi( W x_i ) ) ^2 $,
where $\phi$ is a sigmoid function. 
Assume the width $p \geq n$, and there is one index $ k $
such that $x_{i k } \neq x_{j k }$, $\forall i \not=j$. 
Then every local minimum is a global minimum. 
\end{claim}

Unfortunately, it was recently found that the claim did not hold. 
A counter-example to this claim was given in \cite{ding2019spurious}. 
A modification to this claim will be discussed later.

 \textbf{Cavity of the proof of Claim \ref{thm::Yu no local min}.}
To prove Claim \ref{thm::Yu no local min}, \cite{yu95} first proved
the function satisfies the following property (called ``Property PT'' for short).

\begin{definition}(Property PT)\label{def of PT}
We say a function $F $ satisfies Property PT if
starting from any point $\theta$, there exists an arbitrarily small perturbation
 such that from the perturbed point $\hat{\theta}$,  there exists a strictly decreasing path to a global minimum. 
\end{definition}
 
\cite{yu95} claimed that Property PT implies the non-existence of sub-optimal local minimum. This deduction contains a cavity, as demonstrated in Figure \ref{fig4_local_min_nonlinear}\footnote{Although the loss function in Figure \ref{fig4_local_min_nonlinear} does not use sigmoid activation, it does satisfy Property PT. Therefore, it is sufficient to show that ``Property PT does not imply non-existence of sub-optimal local minimum'', implying that the proof approach in \cite{yu95} has a cavity.}.  
If starting from a sub-optimal local-min (red point), after a small perturbation (the blue point), there is a strictly decreasing path (colored in blue) to the global minimum.
Therefore, even if the function satisfies Property PT,  
sub-optimal local minimum can still exist.

\textbf{Existence of sub-optimal local-min for arbitrarily wide networks.}

The cavity of the proof of Claim\ref{thm::Yu no local min} does not imply
the claim itself does not hold, since there may be other proof methods. 
Nevertheless,  \cite{ding2019spurious} proved that Claim \ref{thm::Yu no local min}
 does not hold by providing a counter-example.

  \begin{proposition}
  	\label{cor::deep-bad_localmin}
  	Let $n\geq 3$. For a neural network with sigmoid activation and input data $x_1, \cdots, x_n\in \mathbb{R}$ where $x_i\neq x_j$ for all $i\neq j$, there exists output data $y_1, \cdots, y_n$ such that the empirical loss has a sub-optimal local minimum.
  \end{proposition}
  
 Besides the sigmoid activation, \cite{ding2019spurious} also proved a stronger negative result that for a large class of smooth activation functions, arbitrarily wide and deep networks, generic input data $x_i$'s with dimension $d_x^2+3d_x/2<n$, there exist output data $y_i$'s such that
  sub-optimal local minima exist.
  It is unknown whether allowing picking labels (e.g.
  label smoothing) can eliminate sub-optimal local-min.

\subsection{Absence of bad valleys and basins}
\label{subsec: absence of valleys}

Although sub-optimal local-min can exist for  wide neural networks, researchers indeed found that a large width is critical for good performance.
Thus one may expect that wide networks exhibit some nice geometrical properties.
In this subsection, we review results on such properties.

\textbf{No spurious valley for increasing activations.}

\begin{definition}
  A \emph{spurious valley} is a connected component of a sub-level set
  $  \{ \theta :  F(\theta) \leq c \}$  which does not contain a global minimum of the loss $ F(\theta)$.
\end{definition}

The non-existence of spurious valley guarantees the non-existence of sub-optimal strict local-min. Although there may still exist sub-optimal non-strict local minima, the absence of spurious valley ensures that starting from any of these sub-optimal local-min, there exists a non-decreasing path
(not necessarily strictly decreasing path) to a region with smaller loss \cite{venturi2018neural}.

Reference \cite{venturi2018neural} proved that no spurious valley exists (implying no bad basin) for 1-hidden-layer network with ``low intrinsic dimension''. 
Reference \cite{nguyen2019connected} further proved that there is no spurious valley for wide deep neural networks where the last hidden layer has no less neurons than the number of samples, under a few assumptions on the activation functions. This is given in the following theorem.

\begin{theorem}
Suppose that an arbitrarily deep fully connected neural network $f_\theta(x)$ satisfies the following assumptions:
  \begin{itemize}
    \item The activation function $\sigma$ is strictly monotonic and $\sigma(\mathbb{R}) = \mathbb{R}$;
    \item For any integer $p\geq 2$, there do not exist non-zero coefficients $(\lambda_i, a_i)^p_{i=1}$ with $a_i \not=a_j$, $\forall i\not= j$, such that $\sigma(x)=\sum^p_{i=1}\lambda_i\sigma(x-a_i)$ for every $x\in \mathbb{R}$;
    \item $d_{L -1 } \geq n$;
    \item All the training samples are distinct.
  \end{itemize}
  Then the empirical loss $F(\theta)$ has no spurious valleys.
\end{theorem}%

\textbf{No sub-optimal basin for any continuous activations.} 
The ``no spurious valley'' result in \cite{nguyen2018loss} holds for strictly increasing analytic activation functions, but it does not cover many non-smooth or non-monotone activations that are commonly applied in practice, such as leaky ReLU or swish. Reference \cite{li2018over} analyzed deep over-parameterized neural networks with any continuous activations.
The result relies on a notion called setwise strict local minimum, defined below. 

\begin{definition}[Setwise strict local minimum]
  \label{def_1}
  We say a compact subset $X\in S$ is a strict local minimum of $f:S \rightarrow \mathbb{R}$ in the sense of sets if there exists $\varepsilon>0$ such that for all $x\in X$ and for all $y\in S\setminus X$ satisfying $\|x-y\|_2\leq\varepsilon$, it holds that $f(x)< f(y)$.
\end{definition}

\begin{figure}
  \begin{minipage}[t]{0.5\linewidth}
  \centering
  \includegraphics[width=0.9\linewidth]{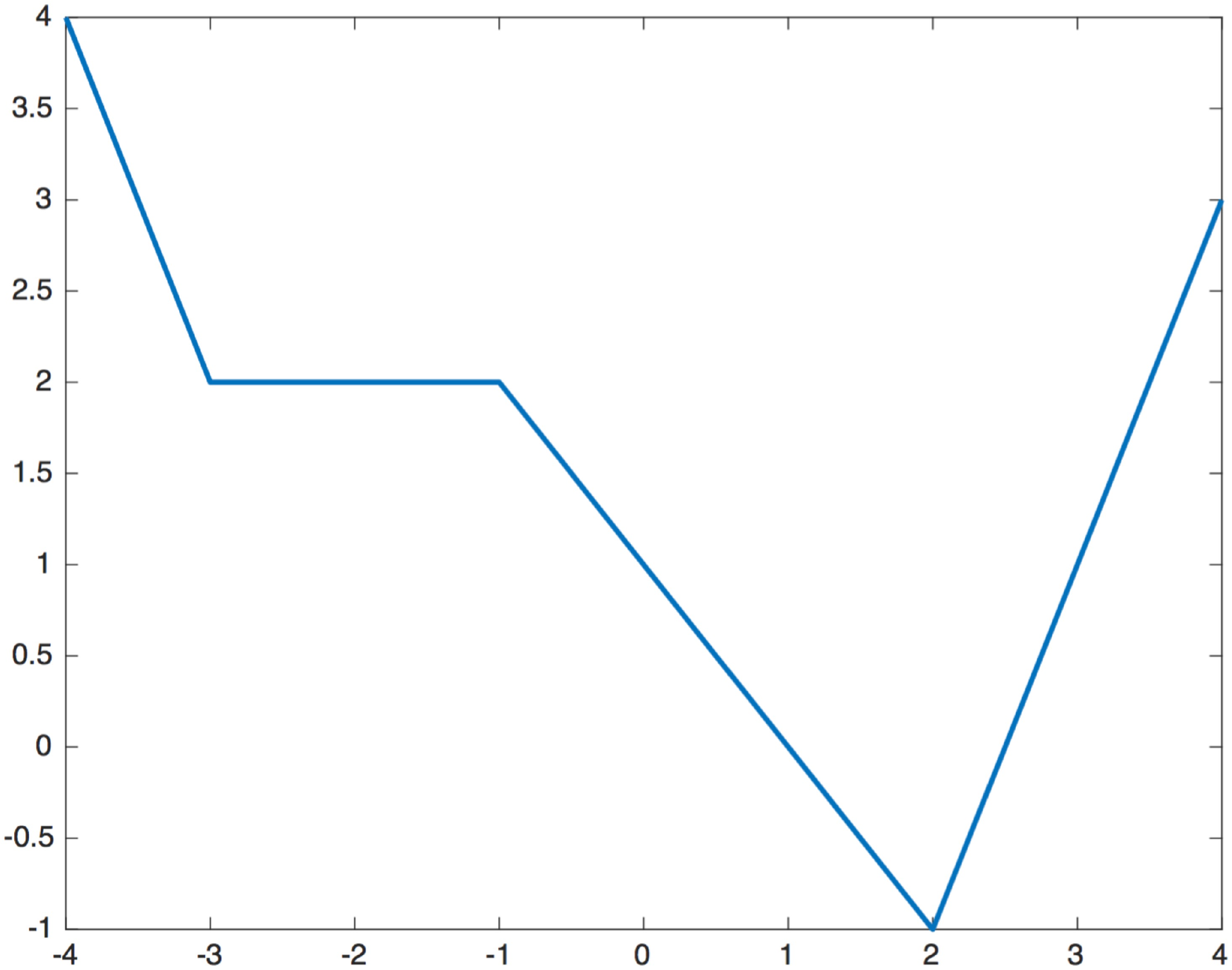}
  \label{fig:side:a}
  \end{minipage}%
  \begin{minipage}[t]{0.5\linewidth}
  \centering
  \includegraphics[width=0.9\linewidth]{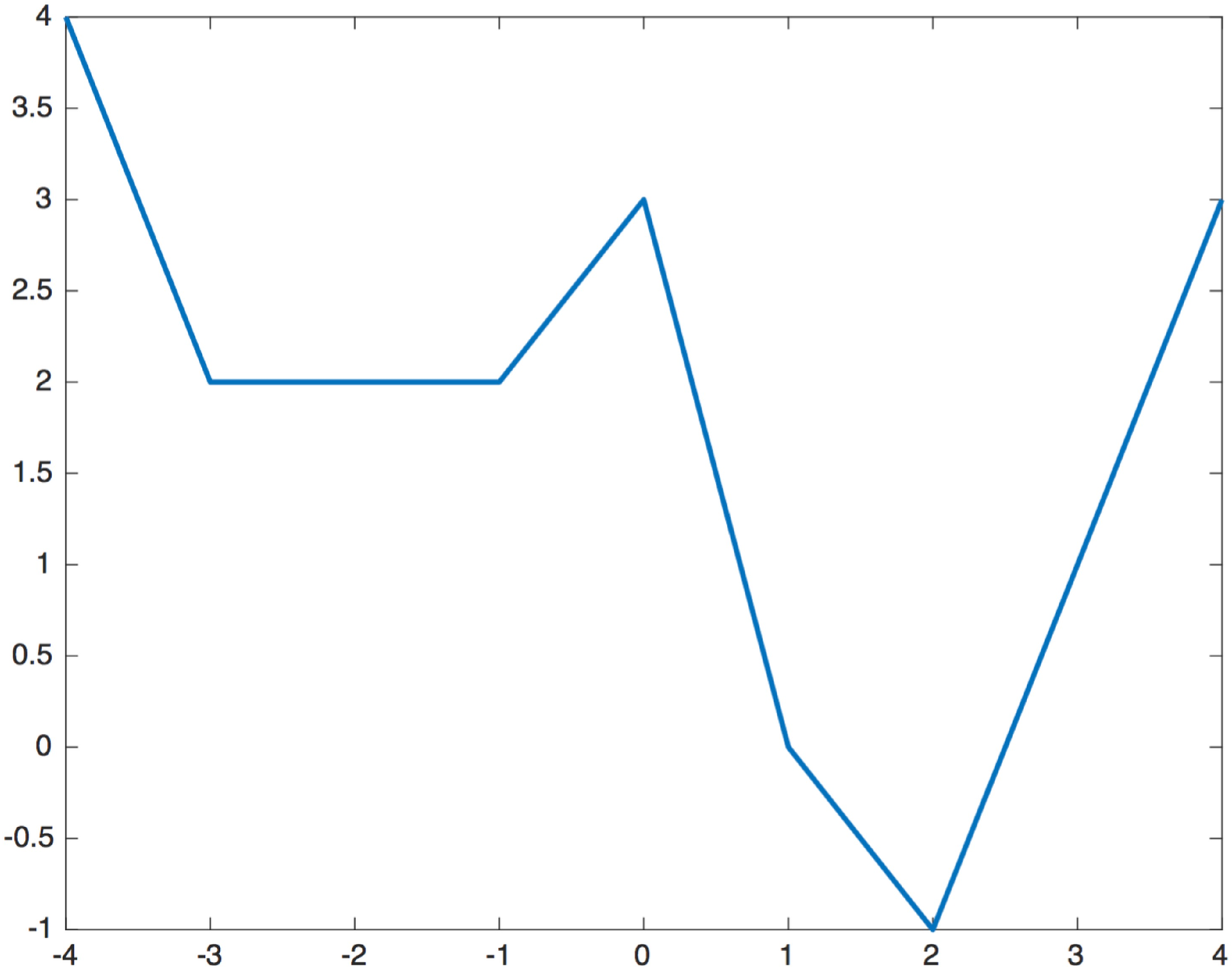}
  \label{fig:side:b}
  \end{minipage}%
  	\caption{%
  	An example of a function without sub-optimal basin (left) and a function with sub-optimal basin (right). Both functions have bad non-strict local minima, consisting a plateau of $(-3, -1)$. Notice that the plateau in the right figure lies in a sub-optimal basin. Source: reprinted from \cite{li2018over}.}
  \label{fig:basin_compare}
\end{figure}

Definition \ref{def_1} generalizes the notion of strict local minimum from the sense of points to the sense of sets. 
Subsequently, we introduce the concept of sub-optimal basin.
\begin{definition}
	\label{def_basin}
A sub-optimal basin of a function $f:S\rightarrow\mathbb{R}$ is a setwise strict local minimum that does not contain a global minimum of $f$.%
\end{definition}

A function that has no sub-optimal basin may still have (pointwise) sup-optimal local minima, which can only form flat areas called ``plateaus''. We note that such plateau cannot lie in a bottom of a sub-optimal basin, as illustrated in Figure \ref{fig:basin_compare}. Reference \cite{li2018over} proved that for all deep neural networks where the last hidden layer is wider than the number of samples, the loss function has no sub-optimal basin.

\begin{theorem}\label{thm: no bad basin}
  Suppose that an arbitrarily deep fully connected neural network $f_{\theta}(x)$ satisfies the following assumptions:
  \begin{itemize}
    \item\label{ass::data} There exists $k$ such that $(x_{i})_k\neq (x_{j})_k, \forall i \not= j $, where
    $ (x_{i})_k $ indicates the $k $-th entry of $x_i $.
		\item\label{ass::overpar} $d_{L - 1 } \geq n$;
		\item\label{ass::neuron} The activation $\sigma_l$ is continuous, $l = 1,\cdots, L$.
  \end{itemize}
Assume the loss function $l(a, b)$ is convex respect to $b$.
  Then the empirical loss $F(\theta) $ defined in \eqref{main problem} has no sub-optimal basin.
\end{theorem}

Remark 1: The two theorems can be generalized to deep
 neural networks with one wide layer (not necessarily the widest);
 see, e.g., \cite[Theorem 2]{li2018over}.
 Due to space, we do not review these results.

Remark 2: ``Sub-optimal basin'' is closely related to ``spurious valley'': every sub-optimal basin must contain a spurious valley, but not vice versa.
If a function has no spurious valley, then it does not have
  set-wise local minima; the reverse is not true \footnote{ 
Not every spurious valley is a sub-optimal basin, because
 a spurious valley is not necessarily compact. 
Not every sub-optimal basin is a spurious valley as well, since the latter has to be a subset of a sub-level set. The simplest statement about their relation
is ``no spurious valley implies no sub-optimal basin''. }. 
Why are there two notions ``valley'' and ``basin''?
Different ``conclusions'' (no spurious valley v.s. no set-wise local minima) require different set of assumptions (strictly increasing smooth neurons v.s. any continuous neuron), thus the two results are currently not  replaceable. It is an open question whether there exists a universal result that includes both results as special cases.

Remark 3: We implicitly assume 
that a global-min exists.
 In machine learning, global-min may not exist and only
 global infimum exists; but for simplicity of presentation,
 we do not add this extra degree of complication
throughout the article.

\subsection{Narrow Networks}

Previous results assume that the network width is large 
(at least $n$). Reference \cite{li2018over} presented a result showing
 $n - 1$ neurons are not enough to eliminate sub-optimal basins.

\begin{proposition}
\label{prop: strict local-min}
    For any $ n $ input data $x_1, \cdots, x_n \in\mathbb{R}$ with 
    $x_i \neq x_j', \forall i \neq j$, there exists $ n $ output  $y_1, \cdots, y_n \in\mathbb{R}$ and a $1$-hidden-layer neural network with $ n - 1$ neurons such that the empirical loss $E(\cdot)$ has bad strict local minimum. 
\end{proposition}

This result together with Theorem \ref{thm: no bad basin}
 demonstrate that adding enough neurons can eliminate sub-optimal basins. 
Note that this result has a number of restrictions (e.g. special output data and special neurons), and a general condition for the existence
of sub-optimal basins requires more research.

\section{Empirical Explorations of Landscape}\label{sec:: empirical exploration}

We have discussed a few theoretical results
on wide neural-nets. 
In this part, we discuss some empirical explorations which reveal non-trivial properties of the landscape
\footnote{Some parts are accompanied with 
theoretical results; anyhow, the main motivation of the whole section is mostly empirical rather than proving theorems.}.
Some of the findings are consistent with the theoretical results we discussed before,
 and some of findings call for more in-depth theoretical understanding.

\subsection{Visualization of  the landscape}

Landscape is a geometrical subject, thus  visualization of the landscape will be very useful for understanding. For one-dimensional or two-dimensional functions, 
it is common to draw the plot $ ( w, f( w) ) $ for $w $ in an interval (1-dim) or 
a box (2-dim), and draw the contour $ \{ w: f(w) = c \}  $ for various
values of $c$.
However, visualizing objects in a high-dimensional space is  difficult in general. A number of dimensionality reduction schemes have been suggested to partially visualize the landscape of neural networks. 

In \cite{goodfellow2014qualitatively}, the authors consider the straight line between two points $ \theta_1  $ and $\theta_2 $, and evaluate the function on the line segment connecting them. 
Consider an algorithm that generates a sequence of points
$ \theta^k = \mathcal{A} (\theta^{k-1}), k=1,2,  \dots, T  $,
where $T$ is the total number of iterations. 
One example is the  gradient descent algorithm  $\mathcal{A}(\theta) = \theta -  \eta \nabla F(\theta ) $ for a certain learning rate $\eta$; instead of GD, \cite{goodfellow2014qualitatively} tested the popular SGD (stochastic gradient descent). 
They pick a random initial point $ \theta^0 = \theta_1 $,
 and pick the converged solution $ \theta^T = \theta_2$, and draw the plot of the function $  F_{ [\theta_1, \theta_2] } ( \alpha ) , \alpha \in [ 0 , 1] $, where
 $$  F_{ [\theta_1, \theta_2] } ( \alpha ) \triangleq 
 F (  \alpha  \theta_1  + (1 - \alpha ) \theta_2  ) .$$
They showed empirically that the function value is decreasing from $\alpha = 0 $ to $\alpha = 1$ (except a small bump near the initial point sometimes). 
This phenomenon will naturally occur when optimizing a convex function, but why this happens in neural network training is largely unknown. We present their finding as the following formal conjecture.

\begin{conjecture}
  Consider a random initial point $\theta^0 $
and suppose SGD generates a sequence $\theta^k, k=1,2, \dots $.
Further, assume the limit $\lim_{k = 1}^{\infty} \theta^k =\theta^* $
 exists. Then under certain conditions on the neural nets, 
 $ F_{ [\theta^0, \theta^*] } ( \alpha ) $ is a strictly decreasing function in the interval $ \alpha \in [0,1]  $. 
\end{conjecture}

\begin{figure}
	\centering
	\subfigure{\includegraphics[width=0.45\linewidth]{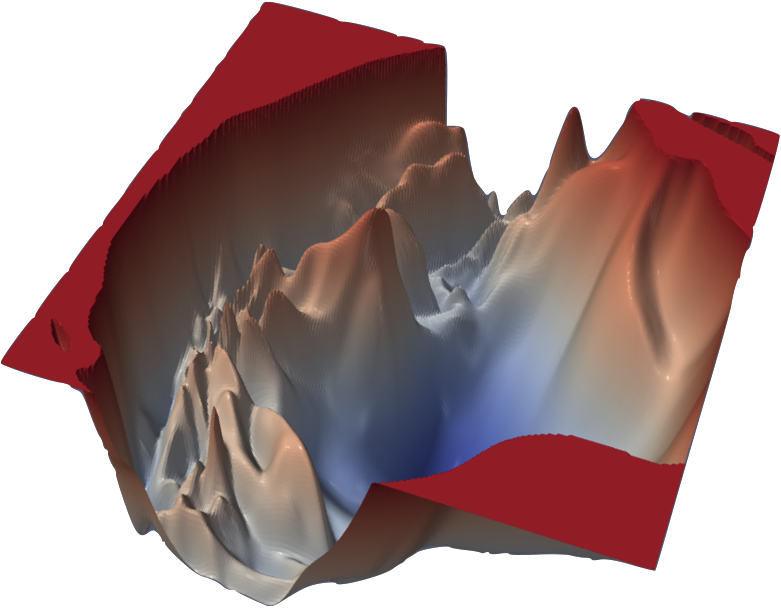}}
    \subfigure{\includegraphics[width=0.45\linewidth]{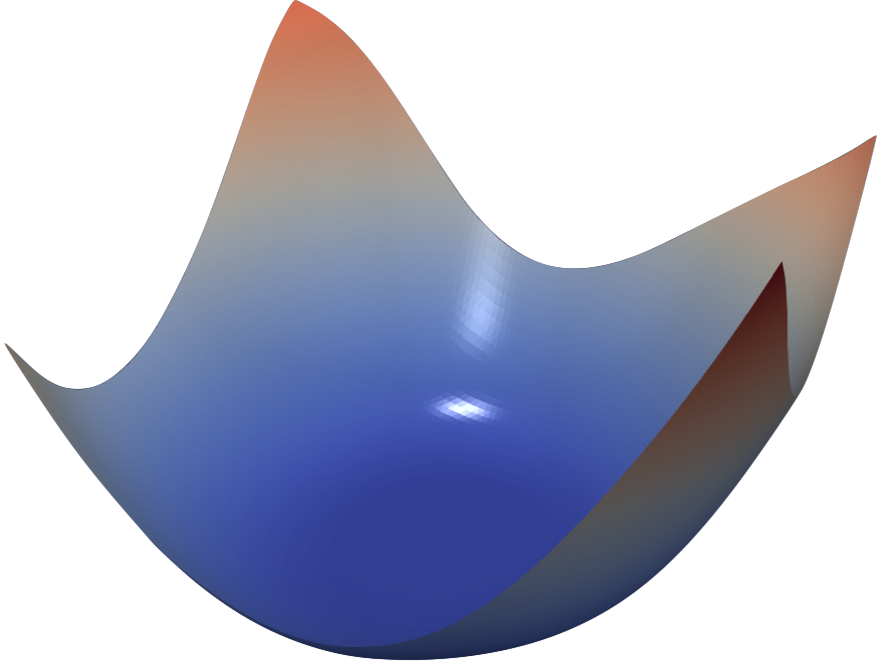}}
	\caption{Figure from \cite{li2018visualizing} visualizing the landscape
		by projecting onto a 2-dimensional space.
		Left: ResNet-110 with no skip connections. Right: DenseNet with 121 layers.
	Source: reprinted from \cite{li2018visualizing}.}
	\label{fig1_2dvisualization}
\end{figure}

Reference \cite{li2018visualizing} visualized  the landscape by projecting it onto a 2-dimensional space.  
More specifically, a center point $ \theta_0 $ and two vectors $v_1$ and 
$v_2$ are picked, and the function values
$
F( \theta_0 +  \alpha v_1 + \beta v_2)  
$
are plotted for $\alpha, \beta \in [  - 1,  1 ]  . $
The basis vectors $v_1, v_2 $ are chosen by a certain special scaling  (called ``filter normalization'') of  random Gaussian vectors. 
It was empirically shown in \cite{li2018visualizing} that the 2-dimensional landscape is highly correlated with the trainability of the networks as demonstrated 
in Figure \ref{fig1_2dvisualization}: deep networks without skip connections are hard to train, and their 2-dimensional landscapes have ``dramatic non-convexities'';
 in contrast, deep ResNet and DenseNet are easy to train in practice and they indeed have convex contours. 
To help readers understand the empirical findings,
we present a conjecture below. 

\begin{conjecture}\label{conjecture: transition}
  Consider a global minimum $\theta^*$,
  and two vectors $v_1, v_2$ drawn by a certain rule (e.g. Gaussian distribution).   Define the function 
  $$  G_{ v_1, v_2 }( \alpha, \beta ) = F( \theta^* +  \alpha v_1 + \beta v_2) , \alpha, \beta \in \mathbb{R} .$$ 
  Then for standard neural nets with width above a threshold $ c_1 $,
  or ResNet with width above a threshold $c_2 < c_1$,
 $ G_{ v_1, v_2 }( \alpha, \beta ) $ has no sub-optimal
 basins. In addition, for standard neural-nets with
 width below a threshold $c_1' < c_1 $, $ G_{ v_1, v_2 }( \alpha, \beta ) $  has many basins. 
\end{conjecture}

Theorem \ref{thm: no bad basin} and Proposition \ref{prop: strict local-min}  discussed earlier (appeared in \cite{li2018over}) show a distinction between narrow and wide networks, thus have a similar flavor to Conjecture \ref{conjecture: transition}.
 Nevertheless, it is unknown whether the original version of Conjecture \ref{conjecture: transition}  can be proved.
 
\begin{figure}
	\centering
	\subfigure{\includegraphics[width=0.23\linewidth]{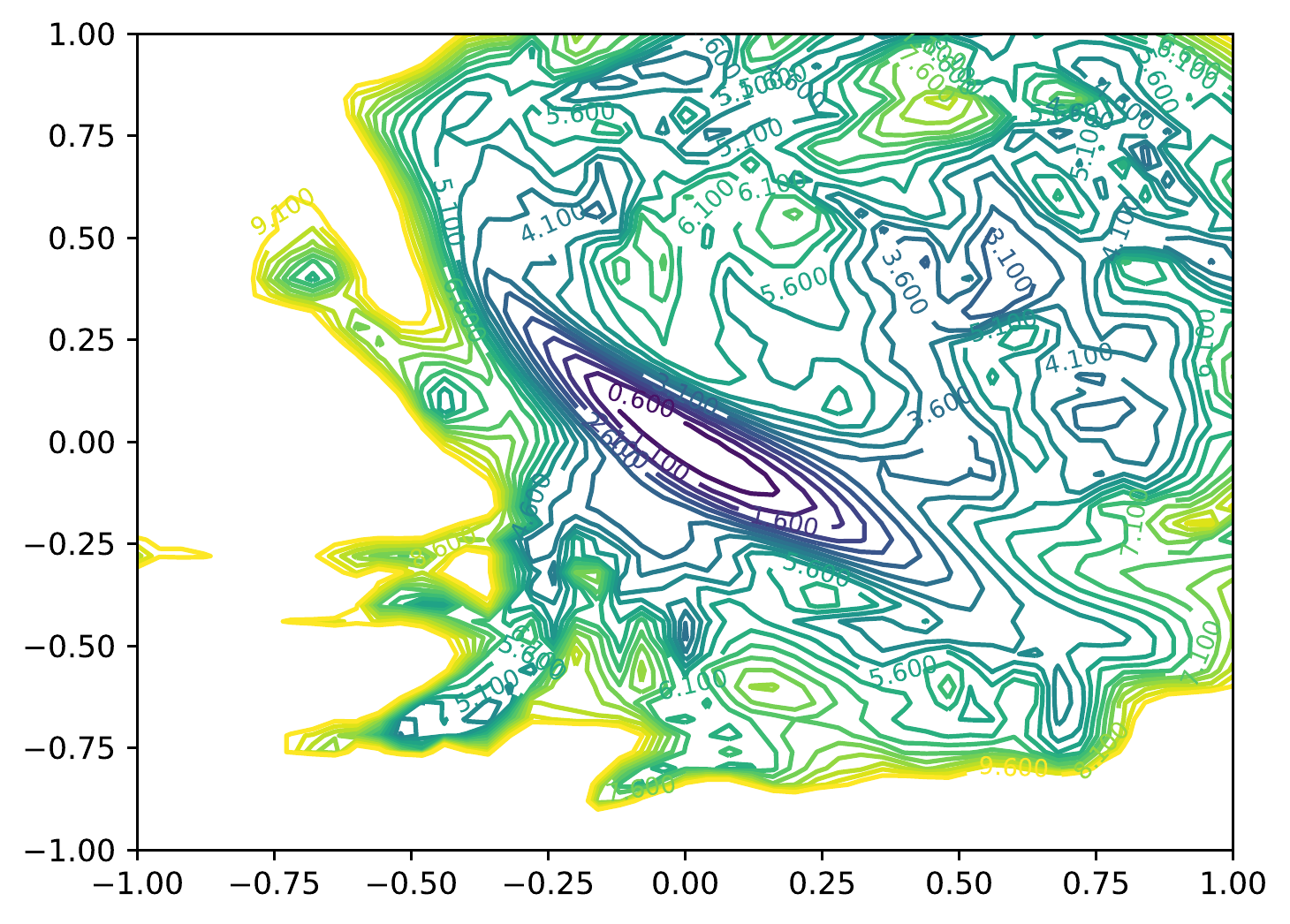}}
\subfigure{\includegraphics[width=0.23\linewidth]{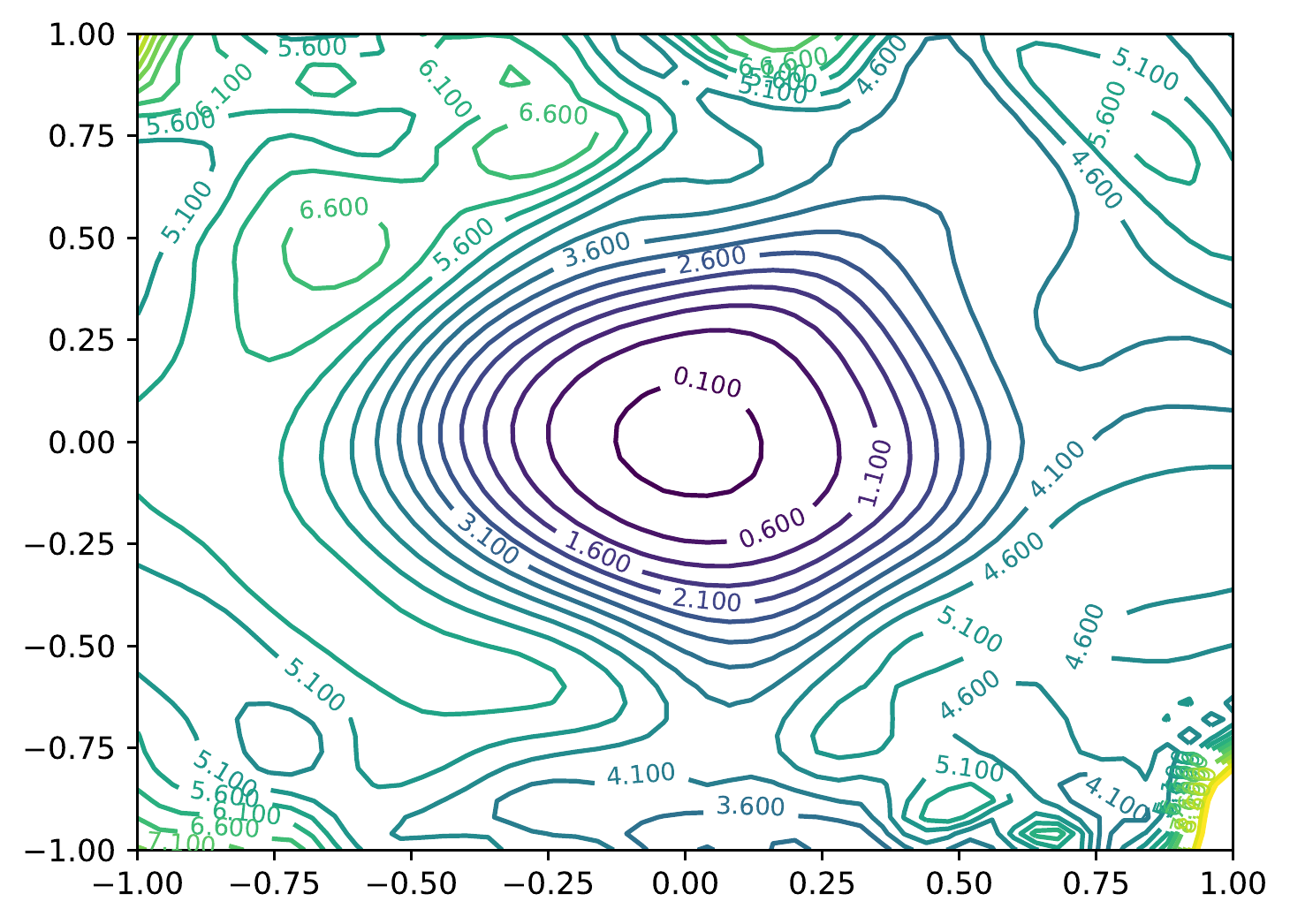}}
\subfigure{\includegraphics[width=0.23\linewidth]{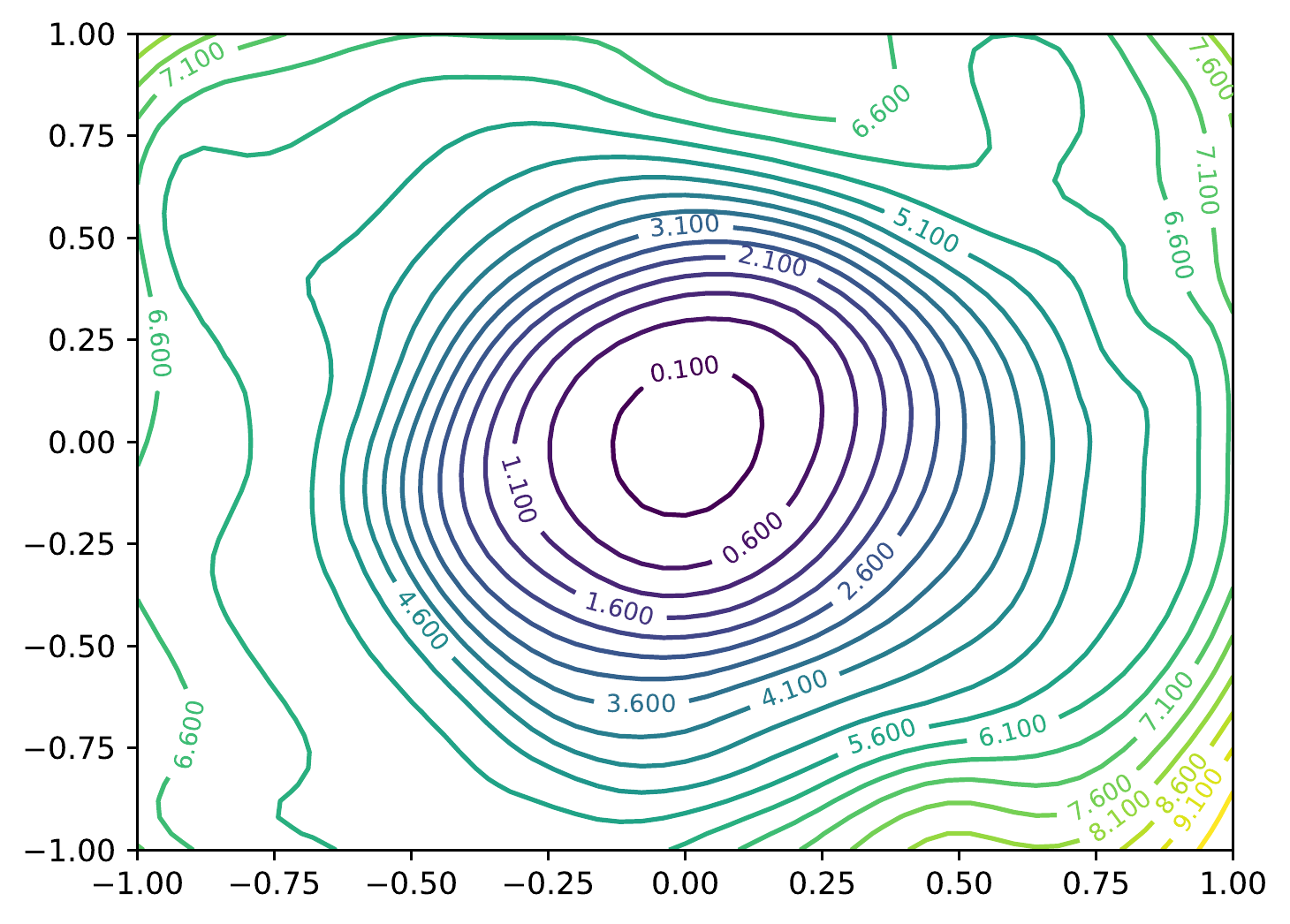}}
\subfigure{\includegraphics[width=0.23\linewidth]{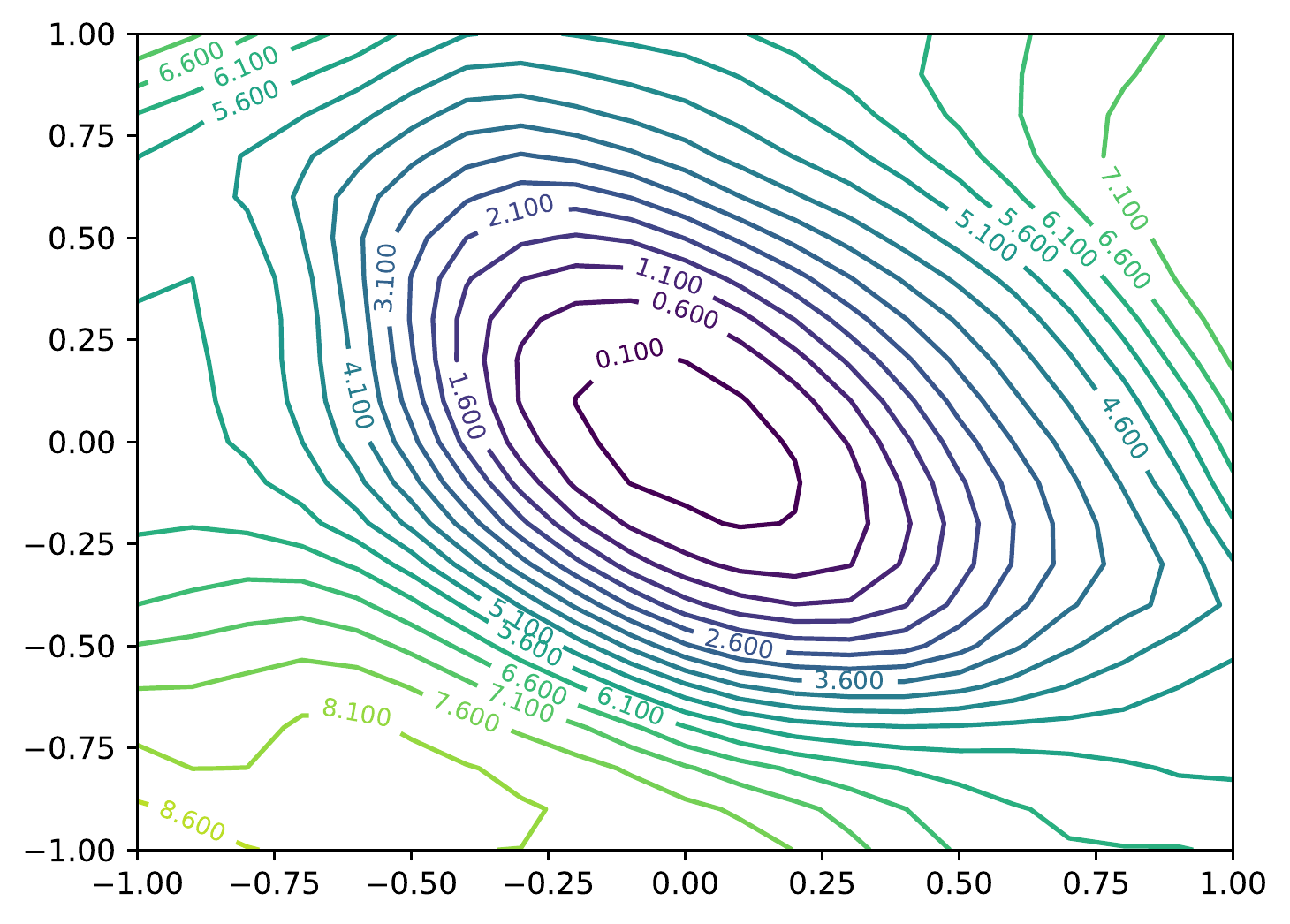}}
	\caption{Figure from \cite{li2018visualizing} visualizing the landscape
		of ResNet-56 with different width.
		 From left to right, the width is 1, 2, 4, 8 times as large as the original ResNet-56.
		 The test error are $ 13.31 \%$, $  10.26 \% $, $ 9.69 \% $, $ 8.70 \%$. 
	Source: reprinted from \cite{li2018visualizing}.}
	\label{fig2_flat}
\end{figure}

\subsection{Mode connectivity}

In this subsection, we present another interesting
empirical finding, supported by some theoretical results. 
 Draxler et al. \cite{draxler2018essentially} and Garipov et al. \cite{garipov2018loss} empirically found that two global minima can be connected by an (almost) equal-value path. This means that the ``modes'' (meaning different global minima) are connected via equal-value path, which explains the terminology ``mode connectivity''.
 We provide a formal description below.

\begin{figure}
	\centering
	\includegraphics[width=8.5cm,height=3.5cm]{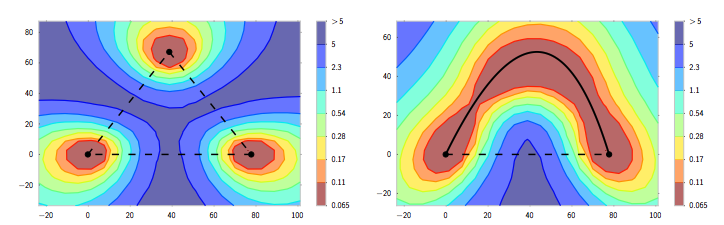}
	\caption{Figure from \cite{garipov2018loss} illustrating mode connectivity. 
		These are the contours of the loss of a 164-layer ResNet trained on CIFAR100,
		as a function of the network weights in a two-dimensional subspace.
		This subspace is spanned by the three points $\theta_1^*, \theta_2^*$ (which are fixed)
		and $\psi$ (which can be changed).
		Here $\theta_1^*, \theta_2^*$ are two solutions (likely local optima) found by training the network from two independent initial points.
		Left figure: $\psi$ is another solution found by training from another initial point,
		and we can see the barriers between the three minima. 
		Right figure:  $\psi$ is found by solving a problem, 
		and we can see a quadratic Bezier curve connecting the two optima along a path of near-constant loss.
	Source: reprinted from \cite{garipov2018loss}.}
	\label{fig1_mode_connect}
\end{figure}

Define $ M(v_1, v_2, v_3) $ as the linear space spanned
by $ v_i, i=1, 2, 3 $ for any three vectors $v_1, v_2, v_3$ (assuming linearly independent). 
Reference \cite{garipov2018loss}
 generated Figure \ref{fig1_mode_connect} (left part)  as follows: first, train a standard 164-layer ResNet 
to find three solutions $\theta_i^*, i=1,2,3$ by starting from three random initial points; second,
define a function $  f( s, t ) = F( s \theta_1 + t \theta_2 + ( 1 - s - t ) \theta_3  ) $, where $ s , t \in \mathbb{R} $; third, draw the contour of the  function $ f(s, t) $ for $ s, t $ in certain intervals. 
Note that we can interpret the three solutions as 
three global minima, even though they are not exact global minima.
The plot shows that the three solutions lie at the bottom of three basins, thus we can make the following conjecture.

\begin{conjecture}\label{conjecture-mode2}
 Suppose  $\theta_i^*, i=1, 2, 3$ are three global minima of $F$. Then in any continuous path in the plane  $ M(\theta_1^*, \theta_2^*, \theta_3^*)  $ that connects $\theta_1^*$ and $\theta_2^* $, the maximum of $ F $ is strictly larger than $F(\theta_1^* ) $. 
\end{conjecture}

 To understand whether $\theta_1^*$ and $\theta_2^*$
 are connected via some equal-value path,
 we can search over the space of paths.
 Figure \ref{fig1_mode_connect} (right part)
 empirically showed  that there exists a simple path
  that connects two global minima. 

\begin{conjecture}\label{conjecture-mode3}
 Suppose $\theta_1^*$ and $\theta_2^* $ are two global minima of $F$. 
There exists $ \theta_0 $ such that the following holds:
there exists a  continuous path in the plane $ M(\theta_1^*, \theta_2^*, \theta_0)  $ that connects $\theta_1^*$ and $\theta_2^* $ and passes $\theta_0$, along which
 the value of $F$ is constant. 
\end{conjecture}

In optimization language, mode connectivity means that
the sub-level set $ \{  \theta \mid F(\theta) \leq F^*  \} $,
which is the same as $ \{  \theta \mid F(\theta) = F^*  \} $,
is connected, where $F^*$ is the global minimal value. 
These findings are partially motivated by 
Freeman and Bruna \cite{freeman2016topology}, who  proved 
a stronger property that the sub-level set
$ \{  \theta \mid F(\theta) \leq c \} $ is connected  for any $c$, for deep linear networks and 1-hidden-layer ultra-wide ReLU networks. 
Kuditipudi et al. \cite{kuditipudi2019explaining} 
and \cite{nguyen2019connected} provided a theoretical justification on this phenomenon; due to space limit, we do not discuss their results in detail.

Now we describe the empirical method used in \cite{draxler2018essentially} and \cite{garipov2018loss} to verify mode connectivity. 
The goal is to  find a equal-value path connecting two global-min.
In practice it is hard to find exact global minima,
thus a reasonable replacement is to train a neural-net 
to find two different solutions $\theta_i^*, i=1,2 $ by starting from two random initial points.
To find a path $P$ connecting two points $\theta_1^*$ and $\theta_2^*$ with ``equal-value'', these works use an optimization problem: find a path with the lowest ``energy'', where the ``energy''  can be defined
in different ways.  \cite{draxler2018essentially} minimizes the ``infinity-norm'' of the path $P$,
i.e.,  solve $ \min_{ P \text{ from } \theta_1^* \text{ to } \theta_2^*  }  \max_{ \theta \in P } F(\theta) $, and \cite{garipov2018loss} minimizes the ``$\ell_1$-norm'' of the path,
i.e., solve $ \min_{ P \text{ from } \theta_1^* \text{ to } \theta_2^*  } 
\mathbb{E}_{ \theta } F(\theta) $, where $\theta$ is drawn from a certain random distribution on the path $P$. 

We briefly discuss the practical tricks used in  \cite{garipov2018loss}. 
There is a huge number of continuous paths from $\theta_1^*$ to $ \theta_2* $.
To restrict the search space of the paths, it considers a subclass of paths, such as the class of Bezier curves  $ \theta_{\psi}(t) = (1-t)^2 \theta^*_1 + 2t(1-t) \psi +  t^2 \theta^*_2, t \in [0,1] $, where $\psi$ is any parameter. Then they use SGD to solve 
$$
\min_{ \psi }  E_{t \sim \text{U}[0,1]  }   F(  \theta_{\psi}(t)  ). 
$$
The result is illustrated in Figure \ref{fig1_mode_connect} (right part). 
The choice of Bezier curve is arbitrary, and
they also report results of using other curves. 

Mode connectivity is not only an interesting geometrical finding,
but has practical implications. For example, mode connectivity implies that once we find two global minima, there is likely to be a connected path between the two minima.
 This provides an opportunity for searching for better minima which yield lower test errors.
  In \cite{garipov2018loss}, such a technique has been proposed.

\subsection{Saddle points or local minima}

An influential, early paper in the recent wave of landscape analysis by Dauphin et al \cite{dauphin2014identifying} studied what points caused training difficulties. It advocated the hypothesis that saddle points instead of local minima are a major issue for neural network training. The underlying logic is the following: under the conjecture that most local minima are close to global minima, if an algorithm gets stuck at a point with a large error, then it is likely to be a saddle point instead of a local minimum. 
This claim is one of motivations for many later works on escaping saddle points (see Sec. \ref{subsec: saddle point}).
Whether saddle points or local minima are a more severe issue remains an interesting question.

\section{Eliminating bad local minima for non-linear networks }

We have discussed that
eliminating sub-optimal local minima globally is  difficult, thus we resort to more complicated concepts such as spurious valleys. In this section, we follow a different path:  we still try to eliminate bad local minima, but allow the modification of other parts of the game.  First, we discuss results that eliminate bad local minima in a subset but not the whole space.
Second, we show how to force all local minima to fall into a subset so that no bad local-min exists. 
 Third, we discuss the limitation of eliminating bad-min, and discuss a stronger landscape property and a result on it.

\subsection{Eliminating bad local-min in a subset}

\textbf{Local minima with full-rank post-activation matrices.}
Reference \cite[Theorem 3.4, 3.8]{nguyen2017} provided conditions for the absence of local-min with certain
 full-rank condition. 
Below we present a different version in \cite{li2018over}.
Define $z_0(x) = x$ and
$ z_k ( x) = \phi( W_{k-1} z_{k-1}(x) + b_k ),
$, $k=1, 2, \dots, L$.
Then we can write $ f_w( x ) =  W_{L}  Z_L( x)  $.
Let $ X = ( x_1, \dots, x_n )$ and let $ Z_L( X ) = ( z_L(x_1), \dots, z_L(x_n)) .  $ 
 
\begin{claim}\label{claim: full rank}
Define  $   \mathcal{W}_{\rm L, full} = \{ \theta: Z_L( X ) \text{ is full rank}  \} $.
  Every local minimum of $F(\theta)$ in the set 
  $ \mathcal{W}_{\rm L, full} $  is a global minimum. 
\end{claim}

\begin{claim}
Suppose the $k$-th order derivative of the activation function $ \sigma^{(k)}(0) $ are non-zero, for $k = 0, 1, 2, \dots, n - 1. $ Then the set $ \mathcal{W}_{\rm L, full} $  is dense. 
\end{claim}

Claim \ref{claim: full rank} is of interest in its own
since it is rather simple. 
The results discussed in Sec. \ref{subsec: absence of valleys} can be viewed as more modern
versions (with no restriction to a subset). 

\textbf{Local minima with full-rank NTK.}
There is another simple result of a similar flavor. 
For simplicity of presentation, we assume $d_y = 1$.
Define 
\begin{equation}\label{NTK half def}
    G(\theta) = ( \frac{ \partial f_{\theta}(x_1)  }{ \partial \theta }, \dots,  \frac{ \partial f_{\theta}(x_n)  }{ \partial \theta } )  \in \mathbb{R}^{ P \times n }
\end{equation}
 where $P$ is the number of parameters,
and define  \textit{neural tangent kernel} (NTK) 
\begin{equation}\label{NTK def} 
K(\theta) = G(\theta)^T G(\theta)  .  
\end{equation}

\begin{claim}\label{claim: full rank NTK}
Suppose $d_y = 1$ and $\ell(a, b) = ( a - b )^2. $
 Define  $ \mathcal{W}_{\rm NTK} =
 \{ \theta: K(\theta) \text{ is full rank}  \} $.
  Every critical point $\theta^*$ of $F(\theta)$ in the set 
  $  \mathcal{W}_{\rm NTK, full} $  is a global minimum
  with zero value. 
\end{claim}

\textbf{Proof:} 
Let $ e_i^* = f_{\theta^* }(x_i) - y_i $
and $ e^* = (e_1^* ;\dots; e_n^* ) \in \mathbb{R}^{n \times 1}$. Since $  F(\theta) = 
\frac{1}{n}\sum_{i=1}^n ( y_i - f_{\theta}(x_i)  )^2 , $
 we have
$$
 \frac{d F(\theta^*) }{ d \theta } 
  = \frac{ 2 }{n} G(\theta^* ) e^*.
$$
If $ \frac{d F (\theta^*) }{ d \theta } = 0  $
and $ G(\theta^*) $ is full rank,
we have $ e^* = 0$ and thus $F(\theta^* ) = 0$,
implying $\theta^*$ is a global-min. 
\textbf{Q.E.D.}

Full-rankness of $ G(\theta) $ is equivalent to
the full-rankness of $K(\theta)$; we do not need $K(\theta)
$ here but we still define $K(\theta)$ since it is critical
in NTK theory.  Despite simplicity, Claim \ref{claim: full rank NTK}  can be viewed as the foundation of the NTK theory we discussed later.

\textbf{Local minima with an inactive neuron.}
The idea of considering local minima with special structure dates back to at least the classical work on Burer-Monteiro factorization \cite{burer2003nonlinear}. 
It showed that for a certain class of non-convex matrix problem, a local-min with a zero column must be a global-min. This result is not directly related to neural nets, but it indicated  an interesting direction. 

Reference  \cite{haeffele2017global} analyzed  a two-layer neural network  $ W_2 \phi(W_1 x )$
with positive homogeneous activations (e.g. ReLU, linear).
It proved that a local-min with one inactive neuron is a global-min (a formal result is somewhat technical and omitted here). Note that the optimization variables are the two weight matrices $W_1 $ and $W_2$,
thus ``an inactive neuron'' means that there is an index $ j $ such that the $j $-th row of $W_1$ and $W_2^{\top }$ are both zero. 
This can be viewed as a variant  of  the result of \cite{burer2003nonlinear}.

\textbf{Comments.}
 Eliminating bad local-min in a subset
itself is of limited interest, since an algorithm
may or may not stay in this subset. 
Extra techniques are required to make these results more interesting.  The three results we presented in this subsection are indeed extended to three stronger results
 ( Theorem \ref{thm: no bad basin},
  NTK theory \cite{jacot2018neural}
  and Theorem \ref{thm::single} respectively). 
 Again, we present them here
 since they are simple and provide some insight. 

\subsection{Making modifications to eliminate bad local-min}
\label{subsec: eliminate local-min}

 \textbf{Eliminating bad local-min by ensuring an inactive neuron.}
 
\begin{figure}
	\centering
	\includegraphics[width=8cm,height=3.5cm]{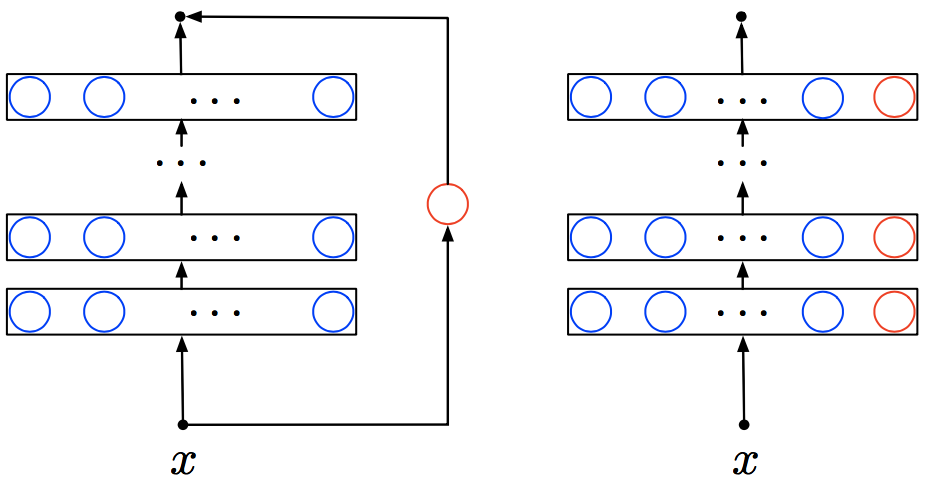}
	\caption{Figure from \cite{liang2018adding} illustrating the modifications.
On the left, a special neuron (e.g. exponential) is added from input to output.
On the right, there architecture is a regular fully connected neural network, and at each layer there is a special neuron (e.g. exponential). 
Source: reprinted from \cite{liang2018adding}.	}
	\label{fig6_special_neuron}
\end{figure}

 \ifarxiv 
 A recent work \cite{liang2018adding} proved that by slightly modifying the neural network  and adding a regularizer, every local minimum is a global minimum.  This can be viewed as an extension of the line of works on local minima with  special sparse structure \cite{burer2003nonlinear,haeffele2017global}
 as a key idea is to force all local minima  to exhibit the special structure. 
 \fi 
 Reference \cite{liang2018adding} provides two modifications
 of the system, each of which can ensure no bad local-min exists, for binary classification. In the first modification, for any deep neural network, \cite{liang2018adding} added a special neuron (e.g. exponential) from input to output and a quadratic regularizer on its weight.
The second modification is to use a special neuron (e.g. exponential) at each layer and add regularizers for the weights connected to these special neurons.
The two modifications are demonstrated in Figure \ref{fig6_special_neuron}.

Below we present the result for the first modification  \cite[Theorem 1]{liang2018adding}. 
 Assume that there exists a $\bm{\theta}$ such that the neural net $f_{\theta}$ can correctly classify all samples in the dataset. 
 Now we add an exponential neuron to the architecture and have a modified function $\tilde{f}(x;\tilde{\bm{\theta}})=
 f_{ \bm{\theta} }(x )+a\exp(\bm{w}^\top x+b)$. For the logistic loss function $\ell(y, z)=\log_2(1+e^{-yz})$, 
 we consider a modified loss function
\begin{equation}\label{eq::loss-single}
\tilde{L}_{n}(\tilde{\bm{\theta}})=\sum_{i=1}^{n}\ell\left(y_{i}, \tilde{f}(x;\tilde{\bm{\theta}})\right)+\frac{\lambda a^{2}}{2}.\end{equation}
The original loss function is defined as
$$
 L_{n}(\tilde{\bm{\theta}})=
  \sum_{i=1}^{n}\ell\left(y_{i}, 
  f_{\bm{\theta} }(x_i )    )\right).
$$

\begin{theorem}(\cite[Theorem 1]{liang2018adding})
Under the above settings, we have:
\begin{itemize}
\item[(i)]  The  function $\tilde{L}_{n}(\tilde{\bm{\theta}})$ has at least one local minimum. 
\item[(ii)] At every  local minimum, $a = 0$. 
\item[(iii)] Assume that $\tilde{\bm{\theta}}^{*}=(\bm{\theta}^{*},a^{*},\bm{w}^{*}, b^{*})$ is a local minimum of  $\tilde{L}_{n}(\tilde{\bm{\theta}})$, then $\tilde{\bm{\theta}}^{*}$ is a global minimum of $\tilde{L}_{n}(\tilde{\bm{\theta}})$.  Furthermore,  $\bm{\theta}^{*}$ achieves the minimum loss value  on the dataset $\mathcal{D}$, i.e., $\bm{\theta}^{*}\in\arg\min_{\bm{\theta}}{L}_{n}(\bm{\theta})$.
\end{itemize}
\end{theorem}

The proof consists of two steps.
\cite{liang2018adding} first showed that at any critical point of the loss function, the exponential neuron is always inactive. This trick allows us to consider local-min in a subset (the topic of the previous subsection). Then \cite{liang2018adding} proved that a local-min with an inactive neuron is a global-min.

 \textbf{Extension to multi-class classification and regression.}
Reference \cite{kawaguchi2019elimination}  extends \cite{liang2018adding} (the first modification) to the multi-class classification tasks. Similar to the construction proposed by \cite{liang2018adding},  it added an exponential neuron on the output of the neural network for each class and added an $\ell_2$ regularizer for the parameters of all exponential neurons. 
 The  high-level proof ideas adopted  those of \cite{liang2018adding} (though
 with some technical differences).
 It first showed that at every local minimum of the empirical loss function, all exponential neurons are  inactive.  Then it showed that a  local-min with these neurons inactive must be a global-min.

  \textbf{Limitation of eliminating bad local-min.}
  \cite{liang2018adding,kawaguchi2019elimination} showed that it is not difficult to 
prove every local-min is a global-min as long as small modifications can be made.
 However, \cite{kawaguchi2019elimination} argued that there are 
 simple examples where on the modified landscape, there are new  paths leading the original local-min to infinity, and thus a descent algorithm might diverge to infinity. We remark that most works on landscape analysis of neural networks mentioned earlier do not explicitly eliminate the possibility that a descent algorithm will diverge to infinity. 
 For instance, for a 3-layer 1-dimensional linear network  problem $\min_{ x, y, z \in \mathbb{R} } (xyz - 1)^2$,  although  no sub-optimal local-min exists according to \cite{kawaguchi2016deep}, there is a  sequence $(x_k, y_k, z_k) = ( -1/k, \sqrt{k}, 1/k  )  $ diverging to infinity while the function values are decreasing and converging to  $1$, which is clearly a sub-optimal value.  This shows that a ``decreasing path'' to infinity exists even for linear neural networks.

\subsection{Eliminating bad local-min and decreasing path to infinity}
 \label{subsec: add reg to smooth}

A natural question is then whether one can further eliminate the possibility of decreasing path to infinity as well as sub-optimal local-min. 
All the results we discussed so far cannot satisfy both properties together.
 Some results prove no sub-optimal local-min \cite{kawaguchi2016deep, liang2018adding,liang2018understanding,kawaguchi2019elimination}, but their loss functions may have decreasing path to infinity. 

Reference ~\cite{liang2019revisiting} 
provides a positive answer to the question.  \cite{liang2019revisiting} considers over-parameterized neural-nets with arbitrary depth. 
For simplicity of presentation, we state their result for
a 1-hidden-layer network.
This network can be expressed by 
$f(x;\bm{\theta})=\sum_{j=1}^{m}a_{j}\requ \left(\bm{w}_{j}^{\top}x+b_{j}\right),$
where the scalar $a_{j}$,  vector $\bm{w}_{j}$, scalar $b_{j}$ denote the coefficient, weight vector,  bias of the $j$-th neuron and $w_j$'s) in the neural network and the activation function is $\text{ReQU}(z)=[\max\{z,0\}]^{2}$. 
Suppose the loss function is logistic: $\ell(y ;  z)=\log(1+e^{- y z} )$. %
Furthermore, reference~\cite{liang2019revisiting} assumes that the data points $x_1, \dots, x_n$ are distinct.  %
The loss to minimize is
\begin{equation}\label{eq::loss-single}
\begin{split}
F(\bm{\theta}) & =\sum_{i=1}^{n}\ell( y_{i}; f(x_{i};\bm{\theta}))+  \\
  & \frac{1}{3}\sum_{j=1}^{m}\lambda_{j}\left[|a_{j}|^{3}+2\left(\|\bm{w}_{j}\|^{2}_{2}+b_{j}^{2}\right)^{3/2}\right],
\end{split}
\end{equation}
where all regularizer coefficients $\lambda_{j}$'s are positive numbers and the vector $\bm{\lambda}=(\lambda_{1},...,\lambda_{m})$ consists of all regularizer coefficients.  \cite{liang2019revisiting} shows   that if the network size is larger than the dataset size, i.e., $m\ge n+1$ and the regularizer coefficient vector $\bm{\lambda}$ is chosen
in a specific way, then every local-min achieves zero training error. 
In this result, we use a standard notion called ``coercive'': we say $F$ is  a coercive function iff $\lim_{ \|  \theta\| \rightarrow \infty  } F(\theta)  = \infty$, thus a coercive function has no decreasing path to infinity. 

\begin{theorem}\label{thm::single}
	Let $m\ge n+1$. There exists a $\lambda_{0}=\lambda_{0}(\mathcal{D},\ell)>0$ and a zero measure set $\mathcal{C}\subset\mathbb{R}^{m}$ such that for any $\bm{\lambda}\in(0,\lambda_{0})^{m}\setminus \mathcal{C}$, both of the following statements are true:
	\begin{itemize}[leftmargin=*]
		\item[(1)] The empirical loss $F(\bm{\theta})$ is coercive.
		\item[(2)] Every local minimum $\bm{\theta}^{*}$ of the loss $F(\bm{\theta}) $ is a global minimum of $F(\bm{\theta}) $,
     and achieves zero training error. %
	\end{itemize}
\end{theorem}

\textbf{Remark: } When all data points are distinct and the size of the ReQU network is larger than the size of the dataset, it is straightforward to show that the every sample in the dataset can be correctly classified by the neural network. In other words, there exists $\theta^*$
such that $F(\theta^*) = 0$. This fact is commonly known as
``over-parameterization implies interpolation''.%

The limitation of the result is that it considers a special neuron called ReQU.
Nevertheless, this result at least shows the possibility of achieving
both ``no bad local-min'' and ``no decreasing path to infinity''.

\section{Algorithmic Analysis}\label{sec: algorithm}

As mentioned in the introduction,
although algorithmic analysis  is very important and closely related to optimization landscape, it is not the focus of this short article.
Nevertheless, we briefly discuss 
convergence analysis for a better big picture.

  \subsection{Intuition: Avoiding Bad Regions}
  \label{subsec: intution and open questions}
 
\begin{figure}
    \centering
    \includegraphics[width = 8cm]{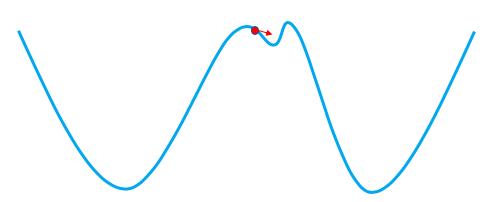}
    \caption{Illustration of a landscape where
bad regions (starting from where
GD will end up at highly sub-optimal values) occupy a small portion.
 Bad regions are hard to escape, but they
 can be avoided for most initial points.
For neural-net problems,
the loss landscape is more complicated than this figure, but we suspect that a similar claim holds:
 bad regions 
 occupy a small portion.
 Results like ``No bad basin or spurious valley''
  provide evidence for this claim, but are
   not enough to prove it. 
  }
    \label{fig:avoiding}
\end{figure}

For a landscape with no bad basin (discussed in Sec. \ref{subsec: absence of valleys}), we expect the
 training is easier than a landscape with bad basins. 
Intuitively, without bad basins,
 global minima are the major attractors for SGD 
(local-min can only attract a tiny subset of points\footnote{
We conjecture that local-min in
wide-neural-net problems are not asymptotically stable
for noisy GD; anyhow, a rigorous result requires further study.}), thus starting from a random initial point the optimization trajectory will converge to global minima
with high probability. 
 See Figure \ref{fig:avoiding} for a conceptual
 illustration of the global landscape: bad regions
 exist, but are rare. We formalize this notion of ``good landscape'' below. 
 
\begin{definition}
 For a given deterministic algorithm $\mathcal{A} $
  that maps a point $ \theta $ to $\mathcal{A} (\theta) $,
  define $ \phi_{\mathcal{A}}( \theta) = \liminf_{k \rightarrow \infty}
 F( \mathcal{A}^{k} (\theta) ) $.
\end{definition}

\begin{conjecture}\label{conj: saddle}(informal)
 Suppose the  neural net has $O(n)$ parameters and the input data are generic. Consider an algorithm $\mathcal{A}(\theta) = \theta - \eta \nabla F(\theta) $.
 For a proper constant $\eta$, and 
 for a random initial point drawn from a certain
 distribution (e.g. Xavier initialization),
   $ \text{Pr}( \phi_{\mathcal{A}} ( \theta) < \epsilon ) > 1 - \delta,   $ where $\epsilon, \delta$ are certain
   small constants. 
\end{conjecture}

Currently there is a gap between
 the landscape results and the above conjecture.
Although landscape results provide positive evidence
for the conjecture, we may need
 to utilize extra properties of the neural-nets 
 to prove the conjecture.

 Adding noise to GD can further increase
the probability of success, since
 shallow basins can potentially be escaped with noise. 
There is another perspective 
(e.g. \cite{kleinberg2018alternative,zhou2019towards})
on the benefit of noise:
running SGD for a function can be seen as running GD for a ``smoothed function''\footnote{We remark that this perspective appeared in other context as well; for instance, \cite{nesterov2017random} proved
that a random direction method for a non-smooth function
is equivalent to running another algorithm on a Gaussian-smoothed version of the original function. }. 
Shallow local-min of the original function may disappear in the ``smoothed landscape''.
Nevertheless, \cite{kleinberg2018alternative,zhou2019towards} only studied special shallow networks. A general analysis of deep neural-nets using this perspective is still missing.

 \subsection{Escaping Bad Points}
 \label{subsec: saddle point}

We distinguish two methods for proving
convergence of an algorithm to global-min.
\begin{itemize}
    \item Avoidance-method: for \textit{most} initial points the algorithm can \textit{avoid} bad regions
     along the trajectory, thus converging to global-min.
    \item Escape-method: for \textit{almost all} initial points
 the algorithm can \textit{escape} bad regions,
 thus converging to global-min.
\end{itemize}
 Figure \ref{fig:avoiding} provides
 an example where the avoidance-method can succeed
 but the escape-method fails: there is a positive measure
 of bad region that GD will get stuck, but for most initial points GD avoids this bad region.

 The escape-method  has been used
 in matrix optimization to prove convergence to 
 global-min (see \cite{chi2019nonconvex}).
 A usual pipeline is: first, prove no sub-optimal local-min exists; second, prove every saddle point is a strict
  saddle point (critical point whose Hessian has at least one negative eigenvalue); third, noisy GD converges to a global-min based on general results in, e.g, \cite{lee2016gradient},\cite{jin2017escape}. 
  
  For neural-nets, a theoretical challenge
  is that high-order saddle points (a saddle point that is not a strict saddle point)\footnote{A formal definition
  of high-order saddle points is rather technical
  and is not important for our purpose.
  We refer readers to \cite[Def. 21]{anandkumar2016efficient}.}
 may exist. Reference \cite{anandkumar2016efficient} showed that escaping fourth-order or higher-order saddle points is NP-hard. Therefore, to prove a rigorous convergence result of GD, just proving ``no bad local-min'' is not enough 
 for general non-convex problems.

That being said, for over-parameterized neural-nets,  %
it is possible that the trajectory did not 
pass a saddle point\footnote{Although \cite{dauphin2014identifying}
claimed that convergence to saddle points
might happen for neural-nets, but
the neural-nets they tested are not state-of-the-art 
over-parameterized neural-nets. 
}
since saddle points can only attract
a tiny portion of initial points (as Conjecture
\ref{conj: saddle} states). 
 A more promising approach towards convergence
 to global minima seems to be the avoidance-method. 
 We discuss some results related to
the avoidance-method in the next subsection. 
 
 \subsection{Algorithmic Analysis for Ultra-Wide Networks}
\label{subsec: algorithm convergence}

One way of implementing the avoidance-method is the following two-step approach:
 (a) prove that in a subset of the parameter space every local-min is a global-min;
 (b) prove that under certain conditions the optimization
 trajectory stays in this subset. 

Consider the NTK Gram matrix $K(\theta) = G(\theta)^T G(\theta)  $, where $ G(\theta) = ( \frac{ \partial f_{\theta}(x_1)  }{ \partial \theta }, \dots,  \frac{ \partial f_{\theta}(x_n)  }{ \partial \theta } )  $, assuming $d_y = 1$.

\begin{claim}
Suppose the conditions of Claim \ref{claim: full rank NTK} hold.
Suppose a certain algorithm generates 
a sequence $ \theta(k+1) = \mathcal{A} ( \theta(k) )
, k = 0, 1, 2, \dots , $
where $\mathcal{A}$ is a continuous mapping. 
Assume there exists $ c > 0 $ such that
 $ \lambda_{\min}(G(\theta(k))) \geq c, \forall k $,
 where $ \lambda_{\min}$ indicates the smallest eigenvalue.
Assume a limit point $\theta^*$ of the sequence $\{\theta(k)\}$
is a critical point of $F(\theta)$,
 then $\theta^*$ is a global minimum of $F(\theta)$
 with zero  value.
\end{claim}

\textbf{Proof:}  Since $ \mathcal{A} $
is continuous, the limit point of the sequence
 $\{\theta(k)\}$ also satisfies
 $  \lambda_{\min}(G(\theta^* )) \geq c $.
 According to Claim \ref{claim: full rank NTK},
 $ \theta^* $ is a global-min of $F(\theta)$
 with zero value. \textbf{Q.E.D.}
 
Recent works \cite{jacot2018neural,arora2019exact,lee2019wide} prove that with infinite width (or $ poly(n) $ neurons per layer), $K(\theta)$ stays positive-definite along the trajectory of GD with a random initialization, thus finishing the a key step of the convergence proof.
The power of the NTK framework is not just proving convergence, but also linear convergence rate, and even generalization error bound.
  These aspects are beyond the scope of this article,
 so will not be discussed in detail here. 
 Around the same time as \cite{jacot2018neural},
 references \cite{zou2018convergence,du2018gradient2,allen2019convergence} 
 also prove global convergence of gradient descent under similar ``ultra-wide'' condition; we skip the details of these results here.

Despite the strong conclusions (convergence, convergence rate, etc.), the assumption of a large width in these convergence
results (often a polynomial
 of $n$, at least $\Omega(n^2)$) is not satisfied
  by practical neural nets. Nevertheless, a more important aspect is the theoretical insight. An intuition of NTK theory is that for extremely wide networks, the weights have little change during the whole training procedure, and hence the model behaves as its linearization around the initialization.
However, reference \cite{chizat2019lazy} showed by experiments that the dynamics of the linearized networks is different from the practical training dynamics, thus the existing results based on NTK may not be enough to fully explain the practical training.

When the theory does not fully match practice, we do not
necessarily have to modify theory; we can also modify the practical system (as mentioned in Q3
in the introduction). \cite{jacot2018neural} suggested 
that one could use kernel GD with the kernel being the NTK
to solve machine learning problems. This essentially reduces a complex multi-layer non-linear network to a simple linear model.
We stress that this is a new method, and is different from practical training. \cite{arora2019exact} performed precise computation using kernel GD with NTK, and reported promising results on image classification. 
One possible issue is that kernel GD is much slower than training a neural net due to high dimension (at least for now). 

 A convergence result with all
assumptions being practical (using common gradient descent; arbitrary depth; not too large width; mild data assumption) is still unknown.  
Convergence analysis of neural-nets (including but
not limited to NTK-type analysis) is a very active area of research (besides the aforementioned works, see, e.g., \cite{allen2019convergencernn,arora2019fine,li2018learning, 18minutes,allen2019learning}).
 It probably requires another whole article  to fully review the recent advances. The focus of this article is on the geometric side of neural-nets, as mentioned in the introduction, thus we do not go deeper into  convergence analysis.

\section{Conclusion}
In this article, we reviewed recent progress on the understanding of the global
landscape of neural networks. 
We discussed various empirical  findings on the landscape, and also many theoretical results. 
We first reviewed the results on deep linear networks that no bad local-min exists. We then discussed why a classical claim on ``no bad local-min'' for over-parameterized networks fails to hold, and showed that a more rigorous claim should be no spurious valley (or no bad basin). We discussed how to perturb the loss functions to eliminate bad local-min, the limitation of ``no bad local-min'', and how to obtain a stronger landscape property. Finally, we briefly discussed the existing convergence analysis (especially
NTK). 

While the progress is encouraging, there are still many mysteries on the landscape of neural-nets.
Many questions presented in this article are not answered (search ``Conjecture'' or ``open''
to find them). 
How to leverage the insight obtained from the theory to design better methods/architectures is also an interesting question.

\small
\bibliographystyle{ieee}
\bibliography{ref}

\end{document}